\documentclass{article}

\usepackage[preprint]{colm2026_conference}
\usepackage{fontspec}

\normalfont
\usepackage{microtype}
\usepackage{graphicx}
\usepackage{wrapfig}
\usepackage{trimclip}
\usepackage{xcolor}
\usepackage{booktabs}
\usepackage{array}
\usepackage{colortbl}
\usepackage{float}
\usepackage{tikz}
\usepackage{tcolorbox}
\tcbuselibrary{breakable}
\usepackage{listings}
\usepackage{pgfplots}
\usepackage{amsmath}
\usepackage{amsfonts}
\pgfplotsset{compat=1.18}
\usepgfplotslibrary{groupplots}
\usepackage{hyperref}
\usepackage{url}
\definecolor{abyss}{HTML}{121D36}
\definecolor{polarnight}{HTML}{1A2947}
\definecolor{nebula}{HTML}{2B3F66}
\definecolor{steeltrail}{HTML}{6D87BD}
\definecolor{skytrail}{HTML}{8FA8D8}
\definecolor{starlight}{HTML}{DFE7F5}
\definecolor{warmstar}{HTML}{E8D9C4}
\definecolor{allsparkwordmark}{HTML}{16233F}
\definecolor{allsparkspark}{HTML}{4A659C}
\definecolor{electricblue}{HTML}{3866FF}
\definecolor{covercream}{HTML}{EEF3FA}
\definecolor{coveraccent}{HTML}{3866FF}
\colorlet{pevekpurple}{skytrail}
\colorlet{bargray}{steeltrail}
\colorlet{barlgray}{starlight}

\newfontfamily\outfit[
  BoldFont=Outfit-SemiBold.ttf
]{Outfit-Regular.ttf}

\hypersetup{
  colorlinks=true,
  linkcolor=electricblue,
  citecolor=electricblue,
  urlcolor=coveraccent,
  filecolor=electricblue
}
\setcitestyle{numbers,square,comma,sort&compress}

\usepackage{xeCJK}
\setCJKsansfont[BoldFont=FandolHei-Bold.otf]{FandolHei-Regular.otf}
\setCJKmonofont{FandolFang-Regular.otf}
\usepackage{capt-of}
\usepackage{needspace}
\usepackage{placeins}
\usepackage{tabularx}
\usepackage{longtable}
\usepackage{algorithm}
\usepackage{algpseudocode}
\algrenewcommand{\textproc}{\textnormal}
\algrenewcommand{\algorithmicrequire}{\textbf{Input:}}
\algrenewcommand{\algorithmicensure}{\textbf{Output:}}
\usepackage{tcolorbox}
\tcbuselibrary{breakable}
\usepackage{listings}
\definecolor{s2eCaseInk}{HTML}{1A2947}
\definecolor{s2eCaseBlue}{HTML}{8FA8D8}
\definecolor{s2eCaseLight}{HTML}{EEF2F9}
\definecolor{s2eCaseMuted}{HTML}{536681}
\lstdefinestyle{s2ecasecode}{
  basicstyle=\ttfamily\fontsize{8.1}{10}\selectfont,
  keywordstyle=\bfseries\color{s2eCaseInk},
  commentstyle=\itshape\color{s2eCaseMuted},
  numbers=none, columns=fullflexible, keepspaces=true,
  showstringspaces=false, breaklines=true, breakatwhitespace=false,
  tabsize=2, aboveskip=3pt, belowskip=3pt
}
\newtcolorbox{s2ecasebox}[2][]{
  breakable, title={#2}, title after break={#2 (continued)},
  fonttitle=\bfseries\small, fontupper=\small\raggedright,
  coltitle=s2eCaseInk, colbacktitle=s2eCaseLight,
  colback=white, colframe=s2eCaseBlue!75!white,
  boxrule=0.5pt, arc=3pt,
  left=9pt, right=9pt, top=6pt, bottom=6pt,
  before skip=7pt, after skip=7pt,
  #1
}

\newcommand{\reporttitle}{Capability-Oriented Environment Synthesis from Skills for General Agents}
\title{Skill2Env: \reporttitle}
\author{AllSpark Team}

\begin{document}

\fancyhead{}
\renewcommand{\headrulewidth}{0pt}
\color{abyss}
\thispagestyle{empty}

\vspace*{-0.44in}
\begin{tcolorbox}[
  width=\linewidth,
  colback=covercream,
  colframe=covercream,
  boxrule=0pt,
  arc=14pt,
  outer arc=14pt,
  boxsep=0pt,
  left=20pt,
  right=20pt,
  top=13pt,
  bottom=11pt
]
  {\outfit\fontsize{21.5}{25.5}\selectfont\bfseries\centering
    \textcolor{coveraccent}{Skill2Env:}\hspace{0.25em}\reporttitle\par}
  \vspace{1.45em}
  {\bfseries\centering AllSpark Team\par}

  \vspace{0.75em}
  \begingroup
  \normalfont
  \setlength{\parindent}{0pt}
  \setlength{\parskip}{0pt}
  \renewenvironment{abstract}{}{}
  \begin{abstract}
Executable environments are critical for post-training agents on tasks that require tool use and multi-step interaction, but constructing executable tasks together with their environments remains difficult to scale. Skills provide reusable domain knowledge, operational procedures, and tool-use instructions, but a substantial gap remains between the information contained in a skill and a concrete, challenging task with a complete executable environment. To address this gap, we introduce \textbf{Skill2Env}, a capability-oriented framework that starts from a skill and uses agent capability demands to guide task and environment synthesis. Skill2Env represents these demands through reusable difficulty patterns and instantiates them into task blueprints that specify objectives, challenges, environment facts, information boundaries, and acceptance criteria. These blueprints guide the joint construction of task instructions, execution substrates, workspaces, and rubric-based evaluators around source skills. We further propose \textbf{Iterative Task Hardening}, which uses solver execution evidence to identify insufficiently challenging task designs, strengthen or extend their difficulty-pattern instantiations, and revise the corresponding blueprints and environments. Using 1.5K high-scoring trajectories generated from Skill2Env environments for supervised fine-tuning, we observe consistent improvements across a broad range of agent benchmarks, demonstrating the effectiveness of capability-oriented environment synthesis for agent post-training.
\end{abstract}

  \par
  \endgroup

  \vspace{0.65em}
  \noindent
  \begin{minipage}[b]{0.63\linewidth}
    \outfit\fontsize{8.4}{10.2}\selectfont
    \textbf{Date:} September 27, 2026\\[-0.1em]
    \textbf{Resources:} \href{https://github.com/AllSpark-Research/AgentEnv}{GitHub}
  \end{minipage}%
  \hfill
  \begin{minipage}[b]{0.33\linewidth}
    \raggedleft
    \raisebox{-0.30em}{\includegraphics[height=18pt]{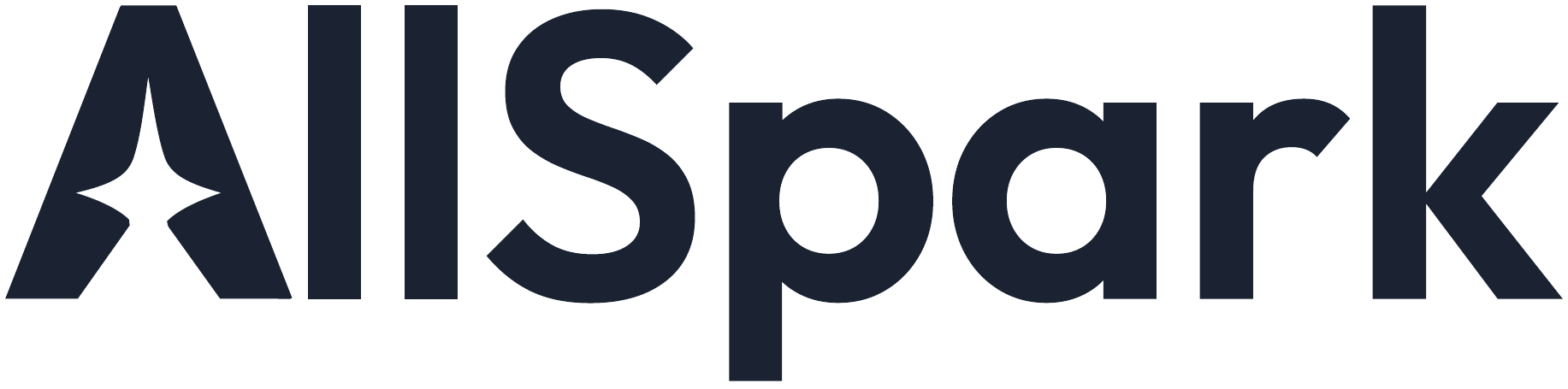}}%
  \end{minipage}
\end{tcolorbox}

\section{Introduction}
\label{sec:introduction}

\begin{figure}[!b]
    \centering
    \includegraphics[width=\linewidth]{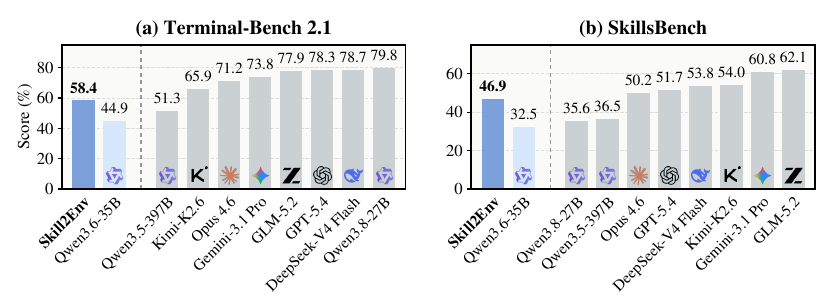}
    \caption{Skill2Env performance on Terminal-Bench 2.1 and SkillsBench.}
    \label{fig:intro-performance}
\end{figure}

As large language models are increasingly deployed in real-world tasks that require tool use and multi-step interaction~\citep{yao2023react,schick2023toolformer}, executable environments have become an important component of post-training for improving agentic capabilities~\citep{pan2025swegym}. Constructing executable tasks and their environments, however, involves more than generating task instructions. It also requires configuring usable tools, preparing a workspace that hosts task resources and state, and establishing evaluation mechanisms aligned with task objectives~\citep{xie2024osworld}. Manually constructing and maintaining these interdependent components on a task-by-task basis is difficult to scale to the volume and diversity required for agent training. Automatically synthesizing executable tasks and their environments is therefore a key approach to scaling agent post-training~\citep{gandhi2026endlessterminals,cheng2026terminalworld}.

Skills provide a starting point for environment synthesis. A skill encapsulates domain knowledge, operational procedures, and tool-use instructions, and may include scripts, templates, and reference resources. Therefore, it can provide useful grounding for task design, tool configuration, and workspace construction. However, a substantial gap remains between the information contained in a skill and the synthesis of a concrete, challenging task and its executable environment. Current work on synthesizing environments from skills has explored several approaches. SkillSynth~\citep{fan2026skillsynth} models agent trajectories as sequences of alternating scenarios and skills; Terminal-World~\citep{cheng2026terminalworld} uses skills to jointly drive task-instruction synthesis and environment construction; FACET~\citep{shi2026facet} ensures consistency through cross-skill scenario reconstruction. Building on these efforts, we ask a further question: given a skill, how can we construct executable tasks and their environments based on agents' capability demands?

To address this question, we propose \textbf{Skill2Env}, a pipeline that starts from a skill and uses capability demands to guide environment synthesis. We structure agents' capability demands along five dimensions: environment understanding, planning, skill usage, long-horizon consistency, and error recovery. We further use difficulty patterns to specify how these demands are translated into concrete environment structures. For each skill, the synthesis agent first selects compatible difficulty patterns and specifies how these patterns should be reflected in a coherent task. It then constructs a task blueprint that jointly specifies the task objective, challenges, environment facts, information boundaries, and acceptance criteria. The blueprint guides the joint construction and refinement of the task instruction, execution substrate, workspace, and rubric-based evaluator around the source skill. For example, a document-processing skill only describes how to process documents using CLI tools, but Skill2Env can turn it into a concrete and challenging reconciliation task that requires the agent to resolve conflicting evidence and maintain consistency across outputs.

We further introduce \textbf{Iterative Task Hardening}, which uses solver
execution evidence to progressively increase task difficulty. After each rollout,
the system identifies aspects of the task that remain insufficiently challenging,
strengthens existing difficulty-pattern instantiations or introduces additional
compatible patterns, and revises the blueprint and environment accordingly.
Reusable challenges revealed during this process can further extend the difficulty
pattern pool. Repeating this process produces increasingly demanding tasks based on
the solver's observed performance.

Our empirical study characterizes the resulting tasks and their environments and evaluates whether the supervision signals obtained from these environments can transfer. To assess transfer, we select 1.5K high-scoring trajectories for supervised fine-tuning and observe improvements in the model across a broad range of agent benchmarks.

Our main contributions are as follows:
\begin{itemize}
    \item We introduce \textbf{Skill2Env}, a capability-oriented framework for skill-based task and environment synthesis. Through difficulty patterns and an explicit task blueprint, Skill2Env connects agent capability demands to the joint construction of task instructions, execution substrates, workspaces, and evaluators around source skills, yielding 2,963 executable tasks.

    \item We propose \textbf{Iterative Task Hardening}, which uses execution evidence from the solver agent to identify insufficiently challenging task designs and progressively strengthen the corresponding capability demands through coordinated updates to task blueprints and environments, yielding more demanding tasks and more effective training supervision.

    \item We demonstrate the effectiveness of Skill2Env for agent post-training. Supervised fine-tuning on 1.5K trajectories generated from Skill2Env environments yields consistent improvements across a broad range of agent benchmarks.
\end{itemize}

\section{Related Work}
\label{sec:related-work}

\paragraph{Executable Environment and Task Synthesis.}
Existing methods for executable environment and task synthesis broadly follow two paradigms. \emph{Workspace-anchored} methods start from an existing or recoverable workspace and derive tasks by restoring, modifying, or perturbing its state, as in repository-, pull-request-, failure-, or trajectory-based synthesis~\citep{jain2025r2egym,chen2026sweuniverse,lin2026cligym,wu2026terminaluniverse}. In contrast, \emph{workspace-from-scratch} methods construct the task-specific workspace from higher-level specifications: Endless Terminals materializes containerized environments from generated terminal-task specifications~\citep{gandhi2026endlessterminals}; CLI-Universe and NexForge construct environments from capability or requirement specifications~\citep{hua2026cliuniverse,zhao2026nexforge}; and skill-based methods synthesize executable environments from reusable agent skills~\citep{fan2026skillsynth,cheng2026terminalworld,shi2026facet,tan2026skt}. Skill2Env follows the latter paradigm and studies how to bridge the gap between the information contained in a skill and the synthesis of a concrete, challenging task and its executable environment.

\paragraph{Skill-Based Environment and Task Synthesis.}
Recent work uses skills to scale the construction of executable training environments and tasks. SkillSynth samples workflow paths from a scenario-mediated skill graph and instantiates them as terminal tasks~\citep{fan2026skillsynth}. Terminal-World jointly derives task instructions, environments, and teacher trajectories from skills, and extends synthesis through skill teams and graphs~\citep{cheng2026terminalworld}. FACET reconstructs coherent scenarios from related skills and uses the realized environment state to ground task instructions, reference solutions, and verifiers~\citep{shi2026facet}. SKT combines rule-based and agent-based verification with feedback-guided repair to generate skill-grounded tasks and successful training trajectories~\citep{tan2026skt}. Building on these efforts, Skill2Env further focuses on how to construct tasks and executable environments based on explicit agent capability demands. It translates these demands into reusable difficulty patterns, and uses execution feedback to progressively harden tasks.

\section{Preliminaries}
\label{sec:preliminaries}

\paragraph{Skills, environments, and tasks.}
An agent skill $\kappa$ is a reusable package of procedural instructions and optional supporting resources for a class of tasks~\citep{li2026skillsbench,xu2026agentskills}. Its instructions encode domain-specific knowledge and procedures and may specify how tools should be used within a workflow. Let $\mathcal X$ denote the execution substrate that provides the tools and runtime support required for execution. A workspace $W$ consists of task-specific local resources and their organization, which the agent can inspect and modify through the available tools. With initial workspace $W_0$, these components define an executable environment and a task instance:
\begin{equation}
    \mathcal E=(\mathcal X,\kappa,W_0), \qquad
    \mathcal T=(g,\mathcal E,V),
    \label{eq:task-instance}
\end{equation}
where $g$ is the task instruction specifying the objective and explicit requirements, and $V$ is a task evaluator. In Skill2Env, environment synthesis is coupled with the generation of a task instruction and evaluator, yielding an executable task instance $\mathcal T$.

\paragraph{Interaction.}
Initializing $\mathcal E$ yields state $s_0$. At step $t$, state $s_t$ includes the current workspace $W_t$ and other runtime state relevant to execution. The agent receives a partial observation $o_t$ and selects an action $a_t$ using policy $\pi$. Given history $h_t=(o_0,a_0,\ldots,a_{t-1},o_t)$ and environment transition mechanism $P_{\mathcal E}$,
\begin{equation}
    a_t\sim\pi(\cdot\mid g,h_t),\qquad
    s_{t+1}\sim P_{\mathcal E}(\cdot\mid s_t,a_t).
    \label{eq:interaction}
\end{equation}
An execution lasting $L$ steps produces a rollout $\tau=(o_0,a_0,\ldots,a_{L-1},o_L)$. After execution, the evaluator assigns a task reward $r=V(s_L)\in[0,1]$.

\section{Skill2Env}
\label{sec:method}

\begin{figure}[t]
    \centering
    \includegraphics[width=\linewidth]{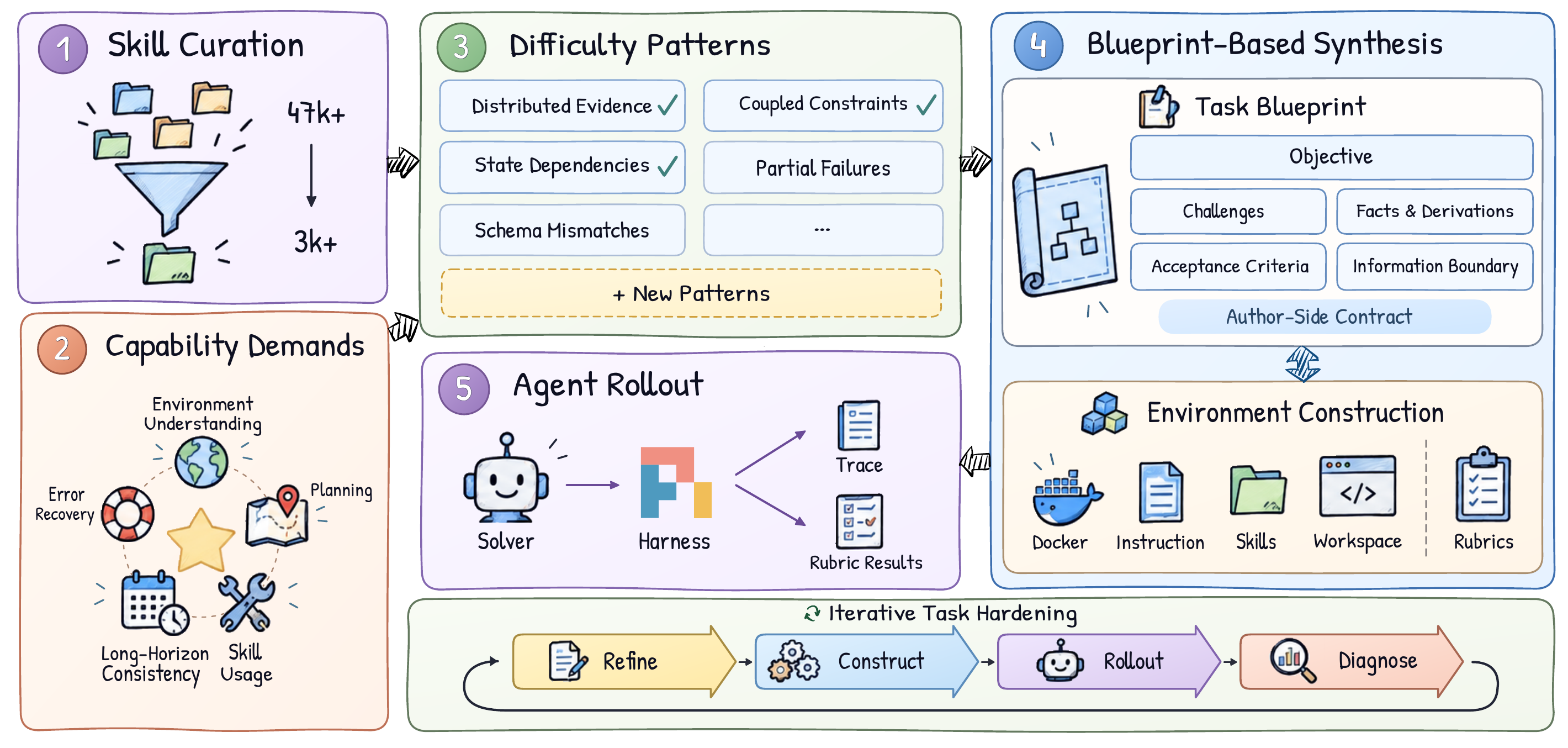}
    \caption{\textbf{Overview of Skill2Env.}
Curated skills and capability demands are connected through difficulty patterns and organized into task blueprints, which guide the construction of executable tasks and their environments.
Solver rollouts provide rubric results, and behavioral evidence for environment validation and task diagnosis.
This feedback further supports difficulty-pattern expansion and Iterative Task Hardening through coordinated updates to task blueprints and environments.}
    \label{fig:skill2env-overview}
\end{figure}

We introduce Skill2Env, a capability-oriented pipeline for jointly synthesizing tasks and executable environments from skills. As shown in Figure~\ref{fig:skill2env-overview}, Skill2Env curates executable skills, operationalizes capability demands through difficulty patterns, and organizes environment construction based on task blueprints. The blueprint serves as an author-side contract for the task instruction, execution substrate, workspace, and rubric-based evaluator. Initial synthesis establishes the intended challenges. Iterative Task Hardening then uses execution evidence to progressively strengthen challenges that the solver already handles well.

\subsection{Skill Curation}
\label{sec:skill-curation}

Skills provide reusable domain knowledge and operational procedures for environment synthesis. We curate skills from public skill libraries that connect this knowledge to practical tool operations, providing both domain context for task design and operational support for execution.

We downloaded more than 47K public skill folders from ClawHub~\citep{openclaw2026clawhub} and screened them using 3 criteria: \textit{Practical Utility}, \textit{Runtime Compatibility}, and \textit{Setup Feasibility}. Practical Utility requires the skill to serve a clear and substantive purpose. Runtime Compatibility requires its hardware, operating system, permission, and compute requirements to be compatible with the target Linux sandbox. Setup Feasibility requires that the inputs, dependencies, data, and services needed for execution can be feasibly provisioned. Applying these criteria, we retained more than 3K skill folders for subsequent environment synthesis.

\subsection{Capability-Oriented Environment Synthesis}
\label{sec:capability-oriented-synthesis}

We next describe how Skill2Env translates agent capability demands into concrete task and environment designs grounded in the curated skills.

\paragraph{From agent capability demands to task difficulty patterns.}
Completing complex tasks typically involves a continuous process of understanding, planning, execution, and adjustment. Throughout the process, an agent needs to continuously interpret task-relevant facts, objectives, constraints, and evolving environment states, and use this evolving understanding to formulate and revise its execution plan. It also needs to interpret and appropriately apply the tool-use instructions and execution procedures provided by the skill according to the current task requirements. As the task progresses across multiple steps, the agent needs to maintain consistency between intermediate results and the environment state. When errors occur, it should use execution feedback to locate problems, adjust its operations, and continue making progress. Based on this execution process, we identify five core capability demands to guide subsequent environment design: environment understanding, planning, skill usage, long-horizon consistency, and error recovery (Table~\ref{tab:capability-dimensions} in Appendix~\ref{app:capability-demands}).

Capability demands specify which agent capabilities a task is intended to challenge. Constructing an executable environment further requires specifying how task design can place demands on these capabilities. We therefore introduce difficulty patterns, each describing a reusable class of task challenges and how to construct them. Each agent capability demand is instantiated through multiple task difficulty patterns, each capturing a distinct source of difficulty. For environment understanding, \emph{distributed evidence} spreads facts across sources, while \emph{implicit constraints} requires inferring unstated rules. For planning, \emph{state dependencies} links later operations to earlier results, while \emph{resource budgets} limits cumulative resource use.

Let $\mathcal P$ denote the difficulty pattern pool. We initialize $\mathcal P$ with 100 reusable patterns, organized into five capability dimensions (Appendix~\ref{app:difficulty-patterns}). The initial patterns are distilled from recurring challenge structures observed during pilot synthesis.

To apply these patterns to a skill $\kappa$, the synthesis agent identifies the workflows it supports and selects a compatible subset $\mathcal P_\kappa \subseteq \mathcal P$. Selection depends on whether these challenges can be naturally integrated into a common task objective and whether the skill supports the operations needed to address them. Different combinations of patterns allow the same skill to support tasks with different capability demands.

\paragraph{Blueprint-guided environment construction.}
Given a skill $\kappa$ and selected difficulty patterns $\mathcal P_\kappa$, the synthesis agent drafts a task blueprint $B$ that instantiates the intended challenges within a coherent task. As an author-side contract, $B$ links task requirements and challenges to supporting facts, acceptance criteria, execution requirements, and information boundaries. Execution requirements specify the necessary tool capabilities, runtime dependencies, and initial resources and state. Information boundaries distinguish what the instruction provides, what the solver must discover in the workspace, and what it must infer. Facts and evidence necessary for solving remain accessible to the solver, while reference outcomes and derivations, where applicable, remain on the author side.

Given the source skill $\kappa$, the synthesis agent uses the blueprint to construct the task instruction $g$, execution substrate $\mathcal X$, initial workspace $W_0$, and rubric-based evaluator $V$, forming the executable task instance $\mathcal T$:
\begin{equation}
    (\kappa,B)\xrightarrow{\text{synthesis}}
    \mathcal T=\bigl(g,(\mathcal X,\kappa,W_0),V\bigr).
    \label{eq:blueprint-synthesis}
\end{equation}
Workspace construction combines synthesized task materials with retrieved real-world materials, organized according to the blueprint's facts and evidence requirements. Blueprint design and environment construction proceed iteratively: the synthesis agent may revise the blueprint to address construction issues or incorporate improved task designs, then update the affected components under the revised contract.

\paragraph{Rubric-based evaluation.}
The evaluator translates the blueprint's acceptance criteria into a set of rubric items $\mathcal C$. Each item $c\in\mathcal C$ is assigned a positive weight $w_c$ during task synthesis, with weights summing to one. Each item yields a score $q_c\in[0,1]$, where $q_c=1$ denotes full satisfaction. Rubric items amenable to programmatic checking use deterministic verification, which typically returns binary scores. The remaining items use an LLM judge, which may assign continuous partial-credit scores according to the item's scoring criteria. Each result is recorded with supporting evidence. The task reward is the weighted average of the rubric results:
\begin{equation}
    r=V(s_L)=\sum_{c\in\mathcal C}w_c q_c(s_L),
    \qquad \sum_{c\in\mathcal C}w_c=1,\quad w_c>0.
    \label{eq:rubric-reward}
\end{equation}

\paragraph{Consistency validation.}
After a solver rollout, a validation agent checks task conditions and rubric results against the blueprint and execution evidence. Issues requiring changes to the task contract are addressed through coordinated updates to the blueprint and affected components. Implementation-only issues are corrected against the existing blueprint. Evidence attributable to inconsistent task conditions or evaluator errors is excluded from capability diagnosis, so task hardening is grounded in valid task demands and rubric results.

\subsection{Iterative Task Hardening}
\label{sec:iterative-task-hardening}

Initial synthesis instantiates the intended capability demands through difficulty patterns, but some resulting challenges may still be readily handled by the solver agent. We therefore introduce \textbf{Iterative Task Hardening}, which uses execution evidence to progressively strengthen task challenges according to the solver's observed performance.

\paragraph{Execution-based diagnosis.}
At iteration $j$, the solver executes task $\mathcal{T}^{(j)}$ constructed from blueprint $B^{(j)}$, producing trajectory $\tau^{(j)}$ and rubric results $\mathbf{q}^{(j)}$. We use this reward as a \emph{hardening trigger}: tasks above a predefined threshold undergo further hardening, while the remaining tasks are retained. For tasks selected for further hardening, the diagnosis agent compares the execution with the challenges specified in the blueprint, examining how the solver handles the instantiated difficulty patterns, whether intended challenges are bypassed through shortcuts, and where residual failures occur:
\begin{equation}
    D^{(j)}
    =
    \operatorname{Diagnose}
    \left(
        B^{(j)},
        \tau^{(j)},
        \mathbf{q}^{(j)}
    \right).
\end{equation}
When diagnosis reveals a generalizable challenge not represented in the current pattern pool, the challenge is abstracted into a reusable difficulty pattern and added to $\mathcal P$ for subsequent task design.

\paragraph{Hardening proposal and task revision.}
Guided by $D^{(j)}$, the system proposes how to strengthen aspects of the task that are already well handled by the solver. The proposal may increase the instantiation strength of existing difficulty patterns or introduce additional patterns from $\mathcal P$ that are compatible with the current skill and workflow. The resulting hardening proposal $H^{(j)}$ is incorporated into the task blueprint, after which the revised blueprint guides coordinated updates to the task instruction, execution substrate, workspace, and evaluator:
\begin{equation}
\begin{aligned}
    H^{(j)}
    &= \operatorname{ProposeHardening}
    \left(\kappa, B^{(j)}, D^{(j)}, \mathcal{P}\right), \\
    B^{(j+1)}
    &= \operatorname{ReviseBlueprint}
    \left(B^{(j)}, H^{(j)}\right), \\
    \mathcal{T}^{(j+1)}
    &= \operatorname{Construct}
    \left(\kappa, B^{(j+1)}\right).
\end{aligned}
\end{equation}

Applying the complete Skill2Env pipeline yields 2,963 executable tasks.
A comprehensive analysis of environment diversity is provided in Appendix~\ref{app:environment-diversity}.

Appendix~\ref{app:docling-case} provides an end-to-end Docling case study illustrating the task blueprint, workspace, evaluator, and successive hardening revisions.

\section{Experiments}
\label{sec:experiments}

We evaluate whether supervision from Skill2Env environments improves agent performance across a range of benchmarks.

\subsection{Experimental Settings}
\label{sec:experimental-settings}

\paragraph{Baselines.}
We compare Skill2Env with frontier foundation models and a prior skill-based environment synthesis method.
The closed-source references are GPT-5.4~\citep{openai2026gpt54}, Claude Opus 4.6~\citep{anthropic2026opus46}, and Gemini-3.1 Pro~\citep{deepmind2026gemini31}.
Open-weight foundation models include DeepSeek-V4-Flash (0731)~\citep{deepseek2026v4}, GLM-5.2~\citep{glm5team2026technical}, Kimi-K2.6~\citep{moonshot2026kimi26}, Qwen3.8-27B~\citep{qwen2026qwen38}, Qwen3.5-397B-A17B~\citep{qwen2026qwen35}, and our backbone, Qwen3.6-35B-A3B~\citep{qwen2026qwen36}.
We include FACET~\citep{shi2026facet} as a skill-based environment synthesis baseline.

\paragraph{Benchmarks and evaluation protocol.}
Our main comparison covers seven benchmarks. Terminal-Bench~2.1~\citep{merrill2026terminalbench} evaluates terminal-task execution; we report task success rate on the complete task set using Terminus-2 with a single trial per task and a 10,800\,s timeout. SWE-bench Multilingual~\citep{yang2025swesmith} evaluates repository-level issue resolution across programming languages; we report resolved-instance rate over all 300 instances using mini-SWE-agent with a single trial per instance and a 7,200\,s timeout. SkillsBench~\citep{li2026skillsbench} evaluates reusable skill use across domains; we report Avg@3 task reward using OpenHands with skills enabled and a 10,800\,s timeout. Claw-Eval~\citep{clawevalrepo} measures autonomous task execution; we evaluate 199 non-multimodal tasks using its native agent loop with three trials per task and report Pass$^3$. $\tau^3$-Banking~\citep{shi2026tauknowledge}, AutomationBench~\citep{automationbench2026}, and VitaBench~\citep{vitabench2025} evaluate knowledge-grounded banking support, cross-application workflows, and interactive service tasks, respectively; we report Pass$^1$, pass rate (version~1.0.6), and mean score. FACET results are source-reported.

\paragraph{Implementation details.}
We use Kimi-K3 for environment synthesis and execution diagnosis, Qwen3.6-35B-A3B as the fixed solver during Iterative Task Hardening, and Qwen3.5-397B-A17B for LLM-based rubric judging. During task hardening, we set the hardening-trigger threshold to $0.7$, such that tasks with reward above this threshold undergo further hardening, while the remaining tasks are retained.

For supervised fine-tuning, the teacher model DeepSeek-V4-Flash generates trajectories on the resulting tasks, from which we retain 1.5K trajectories satisfying the SFT quality filter $r>0.9$. We fine-tune Qwen3.6-35B-A3B on these trajectories for 5 epochs with a global batch size of 32. Training uses Megatron with a peak learning rate of $10^{-5}$, a minimum learning rate of $10^{-6}$, cosine decay, and a 10\% warmup fraction. Unless otherwise specified, the main results are reported using the resulting SFT checkpoint.

\subsection{Main Results}
\label{sec:main-results}

Table~\ref{tab:main-results} shows that Skill2Env improves Qwen3.6-35B-A3B on all seven downstream benchmarks, increasing the unweighted average from 36.6 to 45.0 (+8.4 points) with only 1.5K SFT trajectories. The largest gains are observed on SkillsBench (+14.34) and Terminal-Bench~2.1 (+13.5), while substantial improvements also appear on SWE-bench Multilingual (+7.7), AutomationBench (+7.34), VitaBench (+5.87), Claw-Eval (+5.51), and $\tau^3$-Banking (+4.47). Figure~\ref{fig:intro-performance} highlights the Terminal-Bench~2.1 and SkillsBench results.

\begin{table}[htbp]
    \centering
    \caption{Main results on seven agent benchmarks.}
    \label{tab:main-results}
    \begingroup
    \setlength{\tabcolsep}{2.5pt}
    \renewcommand{\arraystretch}{1.12}
    \fontsize{8}{9.5}\selectfont
    \definecolor{skill2envtableblue}{HTML}{DFE7F5}
    \resizebox{\linewidth}{!}{%
    \begin{tabular}{l*{7}{c}>{\bfseries}c}
        \toprule
        \textbf{Model} & \shortstack{Terminal-\\Bench 2.1}
        & \shortstack{SWE-bench\\Multilingual} & \shortstack{Skills\\Bench} & \shortstack{Claw-\\Eval}
        & \shortstack{$\tau^3$-\\Banking} & \shortstack{Automation\\Bench}
        & VitaBench & Avg. \\
        \midrule
        \rowcolor{black!5}
        \multicolumn{9}{l}{\strut\textbf{Frontier Closed-Source Models}} \\
        GPT-5.4 & 78.3 & 71.7 & 51.7 & 60.3 & 28.5 & 27.7 & 47.1 & 52.2 \\
        Claude Opus 4.6 & 71.2 & 77.8 & 50.2 & 70.4 & 20.3 & 25.5 & 38.3 & 50.5 \\
        Gemini-3.1 Pro & 73.8 & 44.0 & 60.8 & 57.8 & 23.7 & 28.2 & 52.7 & 48.7 \\
        \midrule
        \rowcolor{black!5}
        \multicolumn{9}{l}{\strut\textbf{Open-Weight Models}} \\
        DeepSeek-V4-Flash-0731 & 78.7 & 76.0 & 53.8 & 49.3 & 30.3 & 36.3 & 56.3 & 54.4 \\
        GLM-5.2 & 77.9 & 81.7 & 62.1 & 65.8 & 28.9 & 26.3 & 50.3 & 56.1 \\
        Kimi-K2.6 & 65.9 & 76.7 & 54.0 & 62.3 & 19.9 & 26.3 & 43.9 & 49.9 \\
        Qwen3.5-397B-A17B & 51.3 & 66.0 & 36.5 & 56.8 & 16.2 & 5.5 & 42.1 & 39.2 \\
        Qwen3.6-35B-A3B & 44.9 & 63.3 & 32.5 & 55.8 & 10.7 & 10.3 & 38.9 & 36.6 \\
        Qwen3.8-27B & 79.8 & 73.8 & 35.6 & 69.2 & 33.7 & 38.5 & 41.8 & 53.2 \\
        \midrule
        \rowcolor{black!5}
        \multicolumn{9}{l}{\strut\textbf{Skill-Based Environment Synthesis}} \\
        FACET-Terminal-Qwen3.5-27B & 47.6 & - & - & - & - & - & - & - \\
        \midrule
        \rowcolor{skill2envtableblue!65}
        \textbf{Skill2Env (ours / 35B-A3B)} & 58.4 & 71.0 & 46.9 & 61.3 & 15.1 & 17.7 & 44.8 & 45.0 \\
        \quad \textit{Gain over baseline} & {\scriptsize\color{black}\bfseries (+13.5)} & {\scriptsize\color{black}\bfseries (+7.7)} & {\scriptsize\color{black}\bfseries (+14.3)} & {\scriptsize\color{black}\bfseries (+5.5)} & {\scriptsize\color{black}\bfseries (+4.5)} & {\scriptsize\color{black}\bfseries (+7.3)} & {\scriptsize\color{black}\bfseries (+5.9)} & {\scriptsize\color{black}\bfseries (+8.4)} \\
        \bottomrule
    \end{tabular}%
    }
    \endgroup
    \par\smallskip
    \begin{minipage}{\linewidth}
        \footnotesize\raggedright
        Scores are rounded to one decimal place for display. Reported gains are computed from the unrounded scores before rounding.
    \end{minipage}
\end{table}

Importantly, Skill2Env does not optimize for any particular downstream benchmark. Instead, its synthesized environments are organized around general agent capabilities. The consistent gains across benchmarks with different domains, interfaces, and task structures therefore suggest that supervision constructed around these shared capability demands transfers beyond the synthesized training environments.

Skill2Env also exceeds Qwen3.5-397B-A17B on six of the seven benchmarks, despite using the substantially smaller Qwen3.6-35B-A3B backbone. This result provides additional evidence that the improvement comes from the transferable supervision supplied by the synthesized environments rather than model scale alone. At the same time, frontier-model references remain stronger on several benchmarks, indicating substantial remaining headroom. Overall, the results support the central hypothesis of Skill2Env: executable environments constructed around reusable capability demands can provide supervision that generalizes across heterogeneous downstream agent tasks.

\subsection{Analysis of SkillsBench Improvements}
\label{sec:skillsbench-analysis}

\begin{figure}[t]
    \centering
    \begin{minipage}[b]{0.40\linewidth}
        \centering
        \includegraphics[width=\linewidth]{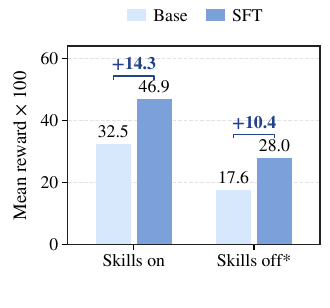}
        \small (a) Overall performance
    \end{minipage}\hfill
    \begin{minipage}[b]{0.58\linewidth}
        \centering
        \includegraphics[width=\linewidth]{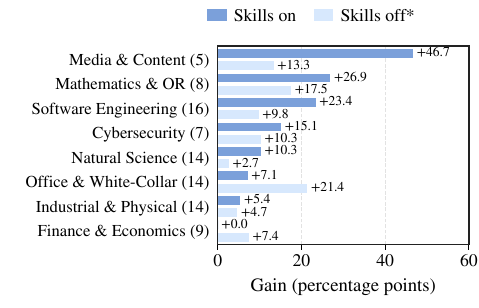}
        \small (b) Gains across domains
    \end{minipage}
    \caption{\textbf{SkillsBench improvements after SFT under OpenHands.}
    (a) Overall scores before and after fine-tuning; blue annotations show absolute gains.
    (b) Absolute score gains across eight domains; parentheses give the number of tasks.
    Scores are mean task reward $\times 100$; gains are differences in percentage points.
    Skills off* disables automatic loading while leaving skill files accessible.}
    \label{fig:skillsbench-transfer}
\end{figure}

Skill2Env constructs environments around five capability demands (Section~\ref{sec:capability-oriented-synthesis}); we examine whether the resulting supervision improves performance on SkillsBench tasks that exercise these capabilities across domains. Figure~\ref{fig:skillsbench-transfer}(a) shows gains of 14.3 percentage points with skills on and 10.4 with skills off\footnote{Skills off disables automatic loading; skill files remain accessible in the environment.} under OpenHands. Gains span seven of eight domains with skills on, with finance unchanged, and all eight with skills off (Figure~\ref{fig:skillsbench-transfer}(b)). Media, mathematics and operations research, and software engineering lead the skills-on gains (+46.7, +26.9, and +23.4 percentage points).

Qualitative trajectory analysis suggests corresponding behavioral improvements. 
In media production, the fine-tuned model better coordinates dependent processing 
stages while maintaining requirements for the final output, reflecting stronger 
planning and long-horizon consistency. In office tasks, it more completely 
identifies and processes content distributed across document structures, reflecting 
improved environment understanding and reducing omissions that prevent task completion. 
Across the examined trajectories, we also observe better adaptation of available 
guidance and helper scripts to task requirements, as well as more effective use of 
execution feedback to diagnose failures, repair operations, and resume progress. 
These behaviors correspond to the five capability demands targeted by Skill2Env and 
suggest that the resulting supervision improves reusable execution capabilities rather 
than benchmark-specific behaviors.

\Needspace{2.6in}
\subsection{Terminal-Bench 2.1 Performance Across Harnesses}
\label{sec:terminal-harness-analysis}

\noindent
\begin{minipage}[t]{0.48\textwidth}
    \vspace{0pt}
We further examine whether the improvements extend across different harnesses. On Terminal-Bench 2.1, the fine-tuned model improves over the base model under all four harnesses (Figure~\ref{fig:terminal-harness-scores}). Scores increase from 46.1 to 57.3 with \texttt{pi} (+11.2 percentage points), from 44.9 to 58.4 with Terminus-2 (+13.5), from 38.2 to 55.1 (+16.9) with OpenCode and from 40.5 to 52.8 (+12.3) with Claude Code harness, respectively. The consistent gains across harnesses support the generalization of agent capability improvements.
\end{minipage}\hfill
\begin{minipage}[t]{0.49\textwidth}
    \vspace{0pt}
    \centering
    \setlength{\parskip}{0pt}
    \includegraphics[width=\linewidth]{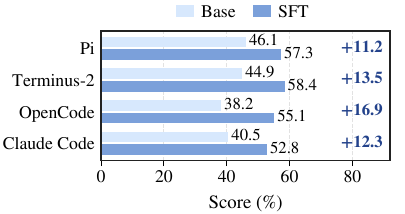}\par
    \captionof{figure}{\textbf{Terminal-Bench 2.1 Performance.} Blue annotations show gains (percentage points).}
    \label{fig:terminal-harness-scores}
\end{minipage}
\par

\subsection{Effectiveness of Iterative Task Hardening}
\label{sec:task-hardening-effectiveness}

We evaluate Iterative Task Hardening from two complementary perspectives: whether successive revisions increase task difficulty and whether the resulting task collections provide more effective training supervision.

\begin{figure}[htbp]
    \centering
    \includegraphics[width=\linewidth]{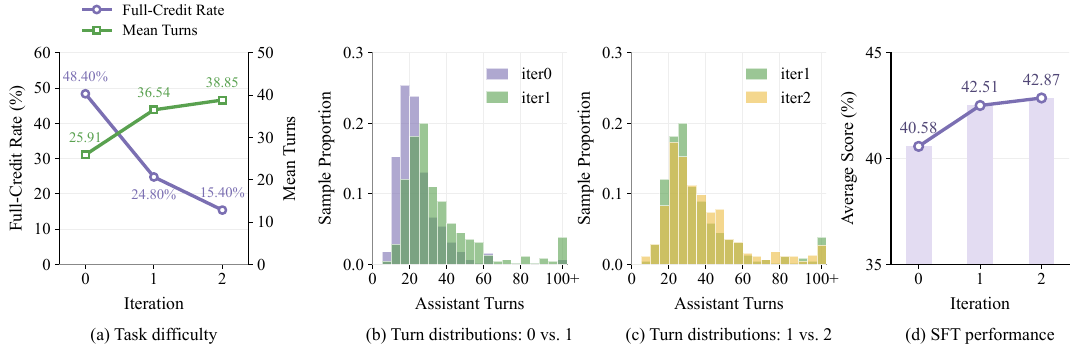}
    \caption{\textbf{Task hardening and downstream training utility.} (a) DeepSeek-V4-Flash full-credit rate ($r=1$, left axis) and mean recorded assistant turns (right axis) over 500 tasks. (b--c) Assistant-turn distributions for 500 tasks at all three iterations, comparing iterations 0--1 and 1--2, respectively. (d) Seven-benchmark mean after SFT on 500 trajectories per iteration.}
    \label{fig:task-hardening-summary}
\end{figure}

\paragraph{Effect on task difficulty.}
To isolate the effect of hardening while controlling for task identity, we use 500 paired task identities that undergo hardening in both rounds and compare their versions at iterations 0, 1, and 2. With DeepSeek-V4-Flash under the \texttt{pi} harness, the full-credit rate ($r=1$) decreases from 48.40\% to 24.80\% and 15.40\%, while the mean number of assistant turns increases from 25.91 to 36.54 and 38.85 (Figure~\ref{fig:task-hardening-summary}(a--c)). From iteration 0 to 2, this corresponds to a 33.0-percentage-point reduction in full-credit rate and approximately 50\% more interaction steps. Together, these paired results indicate that successive hardening revisions make the affected tasks substantially more demanding.

\paragraph{Training utility across hardening stages.}
We next evaluate the training utility of the task collections after zero, one, and two hardening rounds. Tasks that do not undergo further hardening are carried forward unchanged, while hardened tasks are replaced by their revised versions. For each iteration, we apply the same SFT quality filter $r>0.9$ and randomly sample 500 qualifying trajectories, thereby holding both the training-set size and quality threshold fixed. We use random rather than matched samples because the set of qualifying trajectories changes across iterations.

The average score across seven benchmarks increases from 40.58 to 42.51 and 42.87 for iterations 0, 1, and 2, respectively (Figure~\ref{fig:task-hardening-summary}(d)), yielding a 2.29-point improvement from iteration 0 to 2. These results indicate that, under the same data budget and quality threshold, later hardening stages improve average downstream training utility. Appendix~\ref{app:iteration-results} reports the individual benchmark scores.

\section{Conclusion}
\label{sec:conclusion}

We presented Skill2Env, a capability-oriented framework for synthesizing executable tasks and environments from skills. Skill2Env uses reusable difficulty patterns to translate agent capability demands into concrete task challenges, and organizes their realization through task blueprints that coordinate task instructions, execution substrates, workspaces, and evaluators. Iterative Task Hardening further uses execution evidence to progressively revise task designs toward more demanding capability requirements. Across a broad range of agent benchmarks, supervised fine-tuning on 1.5K high-scoring trajectories generated from Skill2Env environments consistently improves agent performance. These results suggest that skill-based environment synthesis can move beyond expanding skill and task coverage by explicitly organizing task construction around the capabilities agents are required to exercise, providing a foundation for generating increasingly demanding environments as agent capabilities improve.

\clearpage
\section{Contributions}
Authors are listed in order of contribution.

\begingroup
\renewcommand{\thefootnote}{\fnsymbol{footnote}}
\noindent
Weiyi Xu\footnotemark[1], Xiaowen Yang\footnotemark[1],
Wen Da\footnotemark[2]\footnotemark[3], Hang Xu, Canwei Li,
Hongjie You, Pusen Dong, Yucheng Zeng, Zhaokai Luo,
Mu Chuan\footnotemark[3]
\footnotetext[1]{Equal contribution.}
\footnotetext[2]{Project lead.}
\footnotetext[3]{Corresponding author.}
\endgroup

\FloatBarrier
\bibliographystyle{styles/pevek/colm2026_conference}
\bibliography{references}

\clearpage
\appendix
\raggedbottom
\section{Capability Demands}
\label{app:capability-demands}

Table~\ref{tab:capability-dimensions} summarizes the five capability demands in Section~\ref{sec:capability-oriented-synthesis}; Appendix~\ref{app:difficulty-patterns} operationalizes them as concrete difficulty patterns.

\begin{table}[H]
    \centering
    \caption{Capability demands used to guide environment synthesis.}
    \label{tab:capability-dimensions}
    \small
    \renewcommand{\arraystretch}{1.15}
    \begin{tabularx}{\linewidth}{@{}p{0.28\linewidth}X@{}}
        \toprule
        \textbf{Capability Demand} & \textbf{Agent Requirements} \\
        \midrule
        Environment Understanding & Identify task-relevant facts from the information provided by the environment and understand the task objectives and constraints. \\ \addlinespace[6pt]
        Planning & Identify dependencies between operations and continually adapt the execution plan based on  evolving environment states. \\ \addlinespace[6pt]
        Skill Usage & Understand and appropriately apply the tool-use instructions and execution procedures provided by the skill according to the current task requirements. \\ \addlinespace[6pt]
        Long-Horizon Consistency & Continuously track state changes during multi-step execution and maintain consistency between intermediate results and the environment state. \\ \addlinespace[6pt]
        Error Recovery & Locate errors using execution feedback, adjust operations, and resume task execution. \\
        \bottomrule
    \end{tabularx}
\end{table}

\section{Difficulty Patterns}
\label{app:difficulty-patterns}
\subsection{Environment Understanding: Patterns}
\label{app:patterns-u}

\begingroup
\small
\setlength{\tabcolsep}{4pt}
\renewcommand{\arraystretch}{1.12}
\setlength{\LTleft}{0pt}
\setlength{\LTright}{0pt}
\setlength{\LTpre}{6pt}
\setlength{\LTpost}{6pt}
\begin{longtable}{@{}>{\raggedright\arraybackslash}p{.22\linewidth}p{\dimexpr.78\linewidth-2\tabcolsep\relax}@{}}
\caption{Difficulty patterns for environment understanding.}\label{tab:patterns-u}\\
\toprule
\textbf{Pattern} & \textbf{Construction and hardening control} \\
\midrule
\endfirsthead
\multicolumn{2}{@{}l@{}}{\tablename~\thetable\ (continued)} \\
\toprule
\textbf{Pattern} & \textbf{Construction and hardening control} \\
\midrule
\endhead
\midrule
\multicolumn{2}{r@{}}{\emph{Continued on next page}} \\
\endfoot
\bottomrule
\endlastfoot
\textbf{U01} Target discrimination & Provide near-identical files or records; relational clues uniquely identify the requested target. \emph{Control: candidate similarity.} \\ \addlinespace[6pt]
\textbf{U02} Distributed evidence & Split complementary facts across known sources so no single source supports the complete answer. \emph{Control: source count.} \\ \addlinespace[6pt]
\textbf{U03} Nested discovery & Place accessible evidence inside nested archives, linked records, or embedded objects. \emph{Control: access depth.} \\ \addlinespace[6pt]
\textbf{U04} Scope enumeration & Require all qualifying objects across partitions or pages, with a discoverable completeness criterion. \emph{Control: partition count.} \\ \addlinespace[6pt]
\textbf{U05} Sparse-signal retrieval & Embed a small set of relevant facts among topically similar but irrelevant passages. \emph{Control: distractor ratio.} \\ \addlinespace[6pt]
\textbf{U06} Authority resolution & Provide conflicting claims with an explicit, accessible hierarchy of authoritative sources. \emph{Control: authority tiers.} \\ \addlinespace[6pt]
\textbf{U07} Temporal validity & Provide individually accurate historical records, only some of which apply to the requested date. \emph{Control: revision count.} \\ \addlinespace[6pt]
\textbf{U08} Provenance reconstruction & Provide copied or aggregated claims whose reliability depends on tracing them to their original source. \emph{Control: attribution depth.} \\ \addlinespace[6pt]
\textbf{U09} Observation completeness & Expose a truncated preview or sampled view; require checking its coverage before interpreting missing items. \emph{Control: omission extent.} \\ \addlinespace[6pt]
\textbf{U10} Measurement reliability & Supply noisy observations with documented error bounds; require a justified estimate rather than trusting one reading. \emph{Control: noise level.} \\ \addlinespace[6pt]
\textbf{U11} Contextual disambiguation & Use an ambiguous request whose intended meaning is uniquely resolved by supplied context. \emph{Control: plausible meanings.} \\ \addlinespace[6pt]
\textbf{U12} Entity resolution & Refer to the same entity using aliases or incompatible identifiers across sources; provide linking evidence. \emph{Control: alias multiplicity.} \\ \addlinespace[6pt]
\textbf{U13} Vocabulary grounding & Use local business terms whose definitions differ from common usage and are documented in the workspace. \emph{Control: definition count.} \\ \addlinespace[6pt]
\textbf{U14} Implicit constraints & Make a rule inferable from accessible configurations or consistent examples without restating it in the instruction. \emph{Control: inference depth.} \\ \addlinespace[6pt]
\textbf{U15} Instruction provenance & Include quoted instructions inside task data; require distinguishing them from the authorized task contract. \emph{Control: distractor salience.} \\ \addlinespace[6pt]
\textbf{U16} Temporal inference & Require event ordering or interval overlap to be inferred from several partial temporal statements. \emph{Control: relation-chain length.} \\ \addlinespace[6pt]
\textbf{U17} Spatial interpretation & Require the intended coordinate frame or orientation to be inferred from local landmarks and calibration evidence. \emph{Control: frame ambiguity.} \\ \addlinespace[6pt]
\textbf{U18} Quantitative inference & Provide observations from which an unstated rate, denominator, or dimensional relationship must be derived. \emph{Control: unknown relations.} \\ \addlinespace[6pt]
\textbf{U19} Causal attribution & Provide controlled comparisons that distinguish causal effects from associations in an otherwise functioning process. \emph{Control: alternative explanations.} \\ \addlinespace[6pt]
\textbf{U20} Negative-evidence reasoning & Require an absence-based conclusion only after establishing that the relevant observation process was complete. \emph{Control: coverage conditions.} \\ \addlinespace[6pt]
\end{longtable}
\endgroup

\subsection{Planning: Patterns}
\label{app:patterns-p}

\begingroup
\small
\setlength{\tabcolsep}{4pt}
\renewcommand{\arraystretch}{1.12}
\setlength{\LTleft}{0pt}
\setlength{\LTright}{0pt}
\setlength{\LTpre}{6pt}
\setlength{\LTpost}{6pt}
\begin{longtable}{@{}>{\raggedright\arraybackslash}p{.22\linewidth}p{\dimexpr.78\linewidth-2\tabcolsep\relax}@{}}
\caption{Difficulty patterns for planning.}\label{tab:patterns-p}\\
\toprule
\textbf{Pattern} & \textbf{Construction and hardening control} \\
\midrule
\endfirsthead
\multicolumn{2}{@{}l@{}}{\tablename~\thetable\ (continued)} \\
\toprule
\textbf{Pattern} & \textbf{Construction and hardening control} \\
\midrule
\endhead
\midrule
\multicolumn{2}{r@{}}{\emph{Continued on next page}} \\
\endfoot
\bottomrule
\endlastfoot
\textbf{P01} State dependencies & Make later actions require specific earlier outputs or state transitions, forming a nontrivial dependency graph. \emph{Control: dependency depth.} \\ \addlinespace[6pt]
\textbf{P02} Bootstrap dependencies & Introduce an apparent dependency cycle with a documented bootstrap action that makes the workflow feasible. \emph{Control: cycle size.} \\ \addlinespace[6pt]
\textbf{P03} Conditional workflows & Make the correct continuation depend on an earlier result, requiring an explicit contingent plan. \emph{Control: branch depth.} \\ \addlinespace[6pt]
\textbf{P04} Subgoal decomposition & Specify one coherent objective that must be decomposed into intermediate milestones and actionable work units. \emph{Control: milestone count.} \\ \addlinespace[6pt]
\textbf{P05} Probe scheduling & Offer several informative probes with different costs; require choosing which uncertainty to resolve next. \emph{Control: probe alternatives.} \\ \addlinespace[6pt]
\textbf{P06} Coupled constraints & Make individually feasible requirements share decision variables, so they must be solved jointly. \emph{Control: coupling density.} \\ \addlinespace[6pt]
\textbf{P07} Priority hierarchies & Specify mandatory obligations and ordered preferences; lower-priority gains cannot override higher-priority requirements. \emph{Control: priority levels.} \\ \addlinespace[6pt]
\textbf{P08} Utility tradeoffs & Provide competing feasible solutions and an explicit utility rule requiring a justified compromise. \emph{Control: objective count.} \\ \addlinespace[6pt]
\textbf{P09} Obligation coverage & Let each candidate action satisfy a subset of requirements; require selecting a set covering all mandatory obligations. \emph{Control: coverage overlap.} \\ \addlinespace[6pt]
\textbf{P10} Optional-goal selection & Offer optional deliverables under a fixed capacity; require selecting which goals to pursue, not merely their order. \emph{Control: optional-goal count.} \\ \addlinespace[6pt]
\textbf{P11} Deadline scheduling & Assign durations and deadlines to dependent actions; require a feasible schedule with limited slack. \emph{Control: deadline slack.} \\ \addlinespace[6pt]
\textbf{P12} Resource budgets & Fix the required goals but limit cumulative expenditure, such as queries, tokens, or simulated cost. \emph{Control: budget tightness.} \\ \addlinespace[6pt]
\textbf{P13} Shared capacity & Make concurrent actions compete for reusable workers, memory slots, or exclusive equipment. \emph{Control: capacity contention.} \\ \addlinespace[6pt]
\textbf{P14} Locality placement & Make execution cost depend on where data and actions are placed; moving inputs incurs a documented cost. \emph{Control: locality asymmetry.} \\ \addlinespace[6pt]
\textbf{P15} Shared-work factoring & Provide related subtasks with reusable intermediate work; independent execution wastes a constrained resource. \emph{Control: reusable overlap.} \\ \addlinespace[6pt]
\textbf{P16} Commitment timing & Include an irreversible but valid action that removes future options; require deciding when enough evidence exists to commit. \emph{Control: lost options.} \\ \addlinespace[6pt]
\textbf{P17} Outcome contingencies & Give actions several legitimate outcomes with known bounds; require reserving feasible continuations before observing the outcome. \emph{Control: outcome uncertainty.} \\ \addlinespace[6pt]
\textbf{P18} Exogenous replanning & Introduce observable changes in legitimate goals or availability, requiring revision of a previously feasible plan without a fault. \emph{Control: change extent.} \\ \addlinespace[6pt]
\textbf{P19} Adaptive stopping & Permit iterative refinement with measurable benefit and cost; require choosing when further refinement is no longer justified. \emph{Control: marginal benefit.} \\ \addlinespace[6pt]
\textbf{P20} Scale-sensitive strategy & Provide workloads where the appropriate task-level strategy changes, such as manual inspection versus indexed or programmatic processing. \emph{Control: workload scale.} \\ \addlinespace[6pt]
\end{longtable}
\endgroup

\subsection{Skill Usage: Patterns}
\label{app:patterns-s}

\begingroup
\small
\setlength{\tabcolsep}{4pt}
\renewcommand{\arraystretch}{1.12}
\setlength{\LTleft}{0pt}
\setlength{\LTright}{0pt}
\setlength{\LTpre}{6pt}
\setlength{\LTpost}{6pt}
\begin{longtable}{@{}>{\raggedright\arraybackslash}p{.22\linewidth}p{\dimexpr.78\linewidth-2\tabcolsep\relax}@{}}
\caption{Difficulty patterns for skill usage.}\label{tab:patterns-s}\\
\toprule
\textbf{Pattern} & \textbf{Construction and hardening control} \\
\midrule
\endfirsthead
\multicolumn{2}{@{}l@{}}{\tablename~\thetable\ (continued)} \\
\toprule
\textbf{Pattern} & \textbf{Construction and hardening control} \\
\midrule
\endhead
\midrule
\multicolumn{2}{r@{}}{\emph{Continued on next page}} \\
\endfoot
\bottomrule
\endlastfoot
\textbf{S01} Capability routing & Provide several documented skill entry points; require selecting the one implementing an already chosen operation. \emph{Control: entry-point similarity.} \\ \addlinespace[6pt]
\textbf{S02} Applicability checks & Make a skill procedure valid only under stated input assumptions that must be checked before applying it. \emph{Control: precondition count.} \\ \addlinespace[6pt]
\textbf{S03} Procedural exceptions & Provide a default procedure with explicit exceptions; some inputs require the exception-specific local handling. \emph{Control: exception density.} \\ \addlinespace[6pt]
\textbf{S04} Example adaptation & Supply a worked skill example whose incidental constants differ from the current case; require transferring the procedure correctly. \emph{Control: example mismatch.} \\ \addlinespace[6pt]
\textbf{S05} Version-specific behavior & Fix a known installed version whose documented operation differs from another version or common example. \emph{Control: version divergence.} \\ \addlinespace[6pt]
\textbf{S06} Parameter binding & Require exact binding of arguments, flags, or option values for a selected operation whose meaning is already known. \emph{Control: parameter coupling.} \\ \addlinespace[6pt]
\textbf{S07} Payload conformance & Require a structured request with nested fields, exact types, and meaningful distinctions between omitted and null values. \emph{Control: schema depth.} \\ \addlinespace[6pt]
\textbf{S08} Invocation context & Require the correct working directory, environment variables, or scoped process context for an otherwise valid invocation. \emph{Control: context conditions.} \\ \addlinespace[6pt]
\textbf{S09} Interaction protocols & Require a documented request-response sequence, such as a prompt handshake or paginated interface, within one operation. \emph{Control: protocol states.} \\ \addlinespace[6pt]
\textbf{S10} Completion protocols & Make an accepted or started operation differ from a completed one; expose a documented readiness or completion signal. \emph{Control: lifecycle stages.} \\ \addlinespace[6pt]
\textbf{S11} Native type fidelity & Require native dates, formulas, identifiers, or typed values; visually similar strings do not preserve their semantics. \emph{Control: type distinctions.} \\ \addlinespace[6pt]
\textbf{S12} Syntax boundaries & Include nested syntax, escaping, or embedded delimiters that require structural parsing rather than naive substitution. \emph{Control: nesting depth.} \\ \addlinespace[6pt]
\textbf{S13} Schema mismatches & Provide valid source and target schemas with different fields, nesting, or grouping; map between them while preserving relationships, values, and units. \emph{Control: structural mismatch.} \\ \addlinespace[6pt]
\textbf{S14} Unit conversion & Provide explicitly known units or coordinate frames; require correct numerical conversion into the target representation. \emph{Control: conversion depth.} \\ \addlinespace[6pt]
\textbf{S15} Precision control & Specify tolerances and rounding points so premature rounding or unstable computation produces an incorrect local result. \emph{Control: precision demand.} \\ \addlinespace[6pt]
\textbf{S16} Rule realization & Provide an explicit domain rule that must be implemented faithfully, including exact thresholds and operator semantics. \emph{Control: rule complexity.} \\ \addlinespace[6pt]
\textbf{S17} Recursive generality & Require a selected procedure to handle arbitrary valid nested compositions rather than only the supplied examples. \emph{Control: composition depth.} \\ \addlinespace[6pt]
\textbf{S18} Boundary handling & Include valid empty, singleton, extreme, or degenerate inputs whose behavior is defined by the procedure contract. \emph{Control: boundary variety.} \\ \addlinespace[6pt]
\textbf{S19} Tie semantics & Require a selected ranking or selection procedure to resolve equal-valued candidates using a documented secondary rule. \emph{Control: tie density.} \\ \addlinespace[6pt]
\textbf{S20} Stochastic reproducibility & Require a randomized procedure to honor specified seeds, sampling semantics, or repeatability conditions. \emph{Control: random-state scope.} \\ \addlinespace[6pt]
\end{longtable}
\endgroup

\subsection{Long-Horizon Consistency: Patterns}
\label{app:patterns-l}

\begingroup
\small
\setlength{\tabcolsep}{4pt}
\renewcommand{\arraystretch}{1.12}
\setlength{\LTleft}{0pt}
\setlength{\LTright}{0pt}
\setlength{\LTpre}{6pt}
\setlength{\LTpost}{6pt}
\begin{longtable}{@{}>{\raggedright\arraybackslash}p{.22\linewidth}p{\dimexpr.78\linewidth-2\tabcolsep\relax}@{}}
\caption{Difficulty patterns for long-horizon consistency.}\label{tab:patterns-l}\\
\toprule
\textbf{Pattern} & \textbf{Construction and hardening control} \\
\midrule
\endfirsthead
\multicolumn{2}{@{}l@{}}{\tablename~\thetable\ (continued)} \\
\toprule
\textbf{Pattern} & \textbf{Construction and hardening control} \\
\midrule
\endhead
\midrule
\multicolumn{2}{r@{}}{\emph{Continued on next page}} \\
\endfoot
\bottomrule
\endlastfoot
\textbf{L01} Identity continuity & Rename or relocate an object during a valid workflow while later steps must continue tracking the same logical object. \emph{Control: identity transitions.} \\ \addlinespace[6pt]
\textbf{L02} Revision coherence & Produce several valid revisions; downstream steps must consistently consume the designated current revision. \emph{Control: revision distance.} \\ \addlinespace[6pt]
\textbf{L03} Derivation lineage & Require intermediate and final artifacts to retain traceable links to the exact source records used to generate them. \emph{Control: lineage depth.} \\ \addlinespace[6pt]
\textbf{L04} Assumption continuity & Choose a convention or mapping early in the workflow and require later steps to preserve that choice. \emph{Control: retention span.} \\ \addlinespace[6pt]
\textbf{L05} Configuration continuity & Run valid steps across sessions or tools whose defaults differ; maintain the intended shared configuration throughout. \emph{Control: context switches.} \\ \addlinespace[6pt]
\textbf{L06} Reference integrity & Update linked objects while preserving valid cross-references, keys, paths, or foreign-key relationships. \emph{Control: reference fan-out.} \\ \addlinespace[6pt]
\textbf{L07} Cross-format agreement & Maintain the same facts in several already-created formats after updates, such as a workbook and a report. \emph{Control: representation count.} \\ \addlinespace[6pt]
\textbf{L08} Implementation agreement & Change implemented behavior and keep its interface documentation, examples, and declared contract synchronized. \emph{Control: coupled artifacts.} \\ \addlinespace[6pt]
\textbf{L09} Aggregate agreement & Revise detailed records and keep all derived totals, summaries, or indexes consistent with the current details. \emph{Control: aggregation levels.} \\ \addlinespace[6pt]
\textbf{L10} Claim-evidence agreement & Maintain consistency between final narrative claims and the actual evidence or results available at completion. \emph{Control: claim dependencies.} \\ \addlinespace[6pt]
\textbf{L11} Non-target preservation & Perform a scoped edit while leaving semantically unrelated content unchanged across the workflow. \emph{Control: protected content.} \\ \addlinespace[6pt]
\textbf{L12} Structural preservation & Change target values while retaining required formatting, nonrelational metadata, or layout across saves and reloads. \emph{Control: protected structures.} \\ \addlinespace[6pt]
\textbf{L13} Monotonic obligations & Satisfy requirements incrementally, but later valid edits can invalidate earlier achievements unless they are rechecked. \emph{Control: interacting obligations.} \\ \addlinespace[6pt]
\textbf{L14} Policy continuity & Maintain explicit privacy, permission, or scope restrictions across intermediate artifacts and final delivery. \emph{Control: propagation paths.} \\ \addlinespace[6pt]
\textbf{L15} Accounting conservation & Maintain conserved quantities or cumulative accounting across otherwise valid updates, including debits, credits, and allocations. \emph{Control: update count.} \\ \addlinespace[6pt]
\textbf{L16} Exactly-once coverage & Process a fixed known worklist across multiple successful steps without omissions or duplicate effects. \emph{Control: worklist length.} \\ \addlinespace[6pt]
\textbf{L17} Deferred obligations & Interrupt a subgoal with other valid work, then return to unresolved obligations rather than forgetting them. \emph{Control: deferral span.} \\ \addlinespace[6pt]
\textbf{L18} Checkpoint fidelity & Persist an intentional checkpoint that faithfully records all state needed for a planned later continuation. \emph{Control: state dimensionality.} \\ \addlinespace[6pt]
\textbf{L19} Interleaving consistency & Allow valid operations to interleave; maintain invariants through ordering, version checks, or isolated updates. \emph{Control: interleaving count.} \\ \addlinespace[6pt]
\textbf{L20} Resource ownership & Track ownership and lifetime of temporary files, processes, locks, or handles across a normally completing workflow. \emph{Control: ownership transfers.} \\ \addlinespace[6pt]
\end{longtable}
\endgroup

\subsection{Error Recovery: Patterns}
\label{app:patterns-r}

\begingroup
\small
\setlength{\tabcolsep}{4pt}
\renewcommand{\arraystretch}{1.12}
\setlength{\LTleft}{0pt}
\setlength{\LTright}{0pt}
\setlength{\LTpre}{6pt}
\setlength{\LTpost}{6pt}
\begin{longtable}{@{}>{\raggedright\arraybackslash}p{.22\linewidth}p{\dimexpr.78\linewidth-2\tabcolsep\relax}@{}}
\caption{Difficulty patterns for error recovery.}\label{tab:patterns-r}\\
\toprule
\textbf{Pattern} & \textbf{Construction and hardening control} \\
\midrule
\endfirsthead
\multicolumn{2}{@{}l@{}}{\tablename~\thetable\ (continued)} \\
\toprule
\textbf{Pattern} & \textbf{Construction and hardening control} \\
\midrule
\endhead
\midrule
\multicolumn{2}{r@{}}{\emph{Continued on next page}} \\
\endfoot
\bottomrule
\endlastfoot
\textbf{R01} Failure recognition & Return a success-like status despite a violated postcondition; expose independent evidence enabling detection of the deviation. \emph{Control: signal discrepancy.} \\ \addlinespace[6pt]
\textbf{R02} Fault localization & Surface an error downstream from its cause; provide logs or checks that identify the faulty component or step. \emph{Control: propagation distance.} \\ \addlinespace[6pt]
\textbf{R03} Cause discrimination & Make several causes produce the same symptom within a localized component; provide probes distinguishing the actual cause. \emph{Control: competing causes.} \\ \addlinespace[6pt]
\textbf{R04} Persistence diagnosis & Provide observable evidence distinguishing a transient interruption from a persistent defect requiring an altered action. \emph{Control: diagnostic delay.} \\ \addlinespace[6pt]
\textbf{R05} Diagnostic preservation & Make inspection or repair alter ephemeral recovery evidence; require preserving a faithful copy before intervening. \emph{Control: evidence lifetime.} \\ \addlinespace[6pt]
\textbf{R06} Input repair & Supply recoverably malformed input with sufficient repair evidence; require restoring validity without inventing missing semantics. \emph{Control: defect extent.} \\ \addlinespace[6pt]
\textbf{R07} Environment repair & Expose a correctable dependency, permission, or configuration fault and a permitted local route to restore execution. \emph{Control: fault interactions.} \\ \addlinespace[6pt]
\textbf{R08} Data restoration & Corrupt mutable task state while retaining a valid backup or redundant evidence from which it can be reconstructed. \emph{Control: recovery distance.} \\ \addlinespace[6pt]
\textbf{R09} Transactional rollback & Interrupt an atomic operation after partial mutation; require restoring its pre-operation state before proceeding. \emph{Control: mutation footprint.} \\ \addlinespace[6pt]
\textbf{R10} Compensating actions & Create partial effects that cannot be rolled back directly; provide allowed counteractions that restore the required business invariant. \emph{Control: compensation chain.} \\ \addlinespace[6pt]
\textbf{R11} Idempotent retries & Lose an acknowledgement after an operation may have committed; require checking its effect before safely retrying. \emph{Control: commit ambiguity.} \\ \addlinespace[6pt]
\textbf{R12} Partial failures & Fail only part of a batch and record per-item outcomes; resume failed work without repeating successful effects. \emph{Control: failure dispersion.} \\ \addlinespace[6pt]
\textbf{R13} Checkpoint recovery & Interrupt execution unexpectedly; resume from the latest valid checkpoint and reconcile work performed after that checkpoint. \emph{Control: lost-work interval.} \\ \addlinespace[6pt]
\textbf{R14} Fallback substitution & Disable a selected implementation after failure while leaving another available implementation that satisfies the same contract. \emph{Control: fallback mismatch.} \\ \addlinespace[6pt]
\textbf{R15} Conflict repair & Expose a detected write or version conflict; require reconciling competing changes before retrying the blocked operation. \emph{Control: conflict scope.} \\ \addlinespace[6pt]
\textbf{R16} Repair validation & Require repeating the original failing check after repair; an apparently successful repair command is insufficient evidence. \emph{Control: failure conditions.} \\ \addlinespace[6pt]
\textbf{R17} Regression validation & Make a repair capable of breaking previously working behavior; require checking the broader preserved contract after repair. \emph{Control: regression surface.} \\ \addlinespace[6pt]
\textbf{R18} Degraded-result qualification & Permit a documented degraded result after failure; require identifying its limitations and meeting the explicit fallback acceptance criteria. \emph{Control: fallback conditions.} \\ \addlinespace[6pt]
\textbf{R19} Cascading repair & Make repairing one defect reveal another previously masked defect, requiring successive diagnosis until the specified postcondition holds. \emph{Control: masking depth.} \\ \addlinespace[6pt]
\textbf{R20} Safe termination & Provide an explicit stop condition when further recovery is unsafe; require bounded termination, preserved state, and an accurate status report. \emph{Control: stop conditions.} \\ \addlinespace[6pt]
\end{longtable}
\endgroup

\section{Environment Diversity}
\label{sec:environment-diversity}
\label{app:environment-diversity}

The synthesis process yields a final collection of 2,963 executable tasks. We analyze environment diversity through source-skill resources, execution substrates, skill-domain coverage, and initial workspaces.

\subsection{Skill Resources}

\begin{figure}[H]
    \centering
    \includegraphics[width=\linewidth]{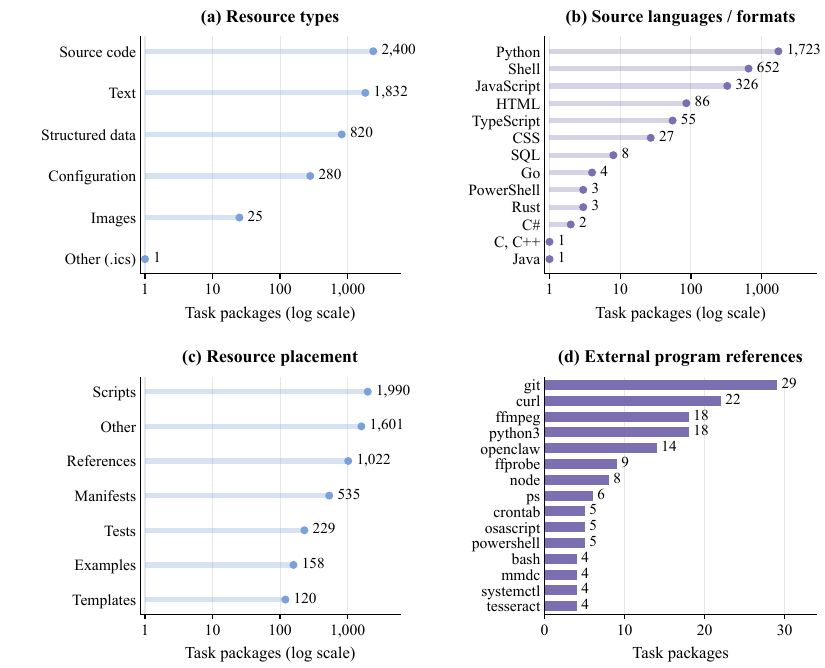}
    \caption{\textbf{Skill-resource diversity across 2,963 task-associated source-skill packages.} (a) Resource types after removing catalog auxiliary files. (b) All 13 source-language or format labels. (c) Resource placement under fixed path and filename rules. (d) The 15 most frequently referenced external programs in the Python subprocess analysis, ranked by the number of packages containing a literal program-name reference; ties are ordered alphabetically. Panels (a)--(c) use logarithmic count axes; (d) uses a linear axis.}
    \label{fig:skill-resource-diversity}
\end{figure}

We first examine the resources contained in the 2,963 source-skill packages.
Across the 2,963 task packages, 2,613 contain auxiliary skill resources beyond \texttt{SKILL.md}, comprising 19,649 files after excluding catalog metadata.
As shown in Figure~\ref{fig:skill-resource-diversity}, these files cover diverse
resource types, including source code, text, structured data, configuration files,
and images. They also span 13 programming or markup languages, with Python and
Shell being the most common, and appear in different package locations such as
scripts, references, tests, examples, and templates.

Beyond static resources, many skills also expose executable interfaces.
Static analysis identifies references to 101 distinct external programs, including
commonly used CLI tools such as \texttt{git} and \texttt{curl}.
Together, these statistics show that the source skills provide heterogeneous
resources, file structures, and executable interfaces for environment construction.

\subsection{Execution Substrate}
\label{app:execution-substrate}
\begin{figure}[H]
    \centering
    \includegraphics[width=\linewidth]{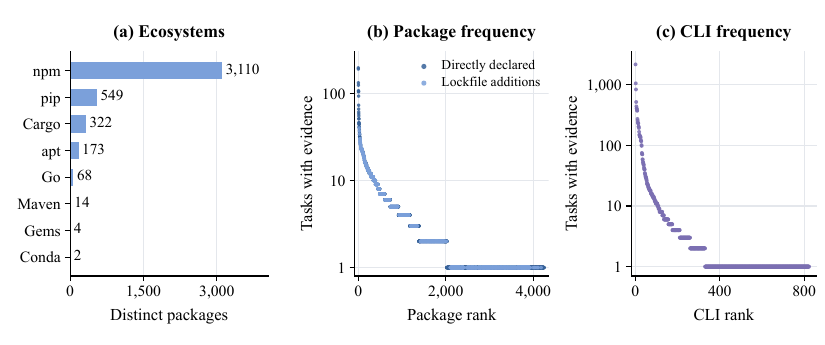}
    \caption{
\textbf{Execution-substrate diversity.}
(a) Number of distinct dependency packages across eight package ecosystems.
(b) Task occurrence frequencies of the 4,242 identified packages, ranked by
frequency.
(c) Task occurrence frequencies of the 823 identified CLI names, ranked by
frequency.}
    \label{fig:execution-substrate-diversity}
\end{figure}

We next examine the software dependencies and command-line interfaces
associated with the 2,963 tasks. Across setup scripts, source-skill packages,
and initial workspaces, we identify 4,242 distinct dependency packages spanning
eight ecosystems and 823 distinct CLI names. As shown in
Figure~\ref{fig:execution-substrate-diversity}, npm contributes the largest
number of package identities, followed by pip and Cargo, while both package
and CLI usage exhibit long-tailed distributions across tasks. These statistics characterize the diversity of the software and executable
interfaces available to the synthesized environments.

\subsection{Skill Domains and Workspaces}
\label{app:skill-domains-workspaces}

\begin{figure}[H]
    \centering
    \includegraphics[width=\linewidth]{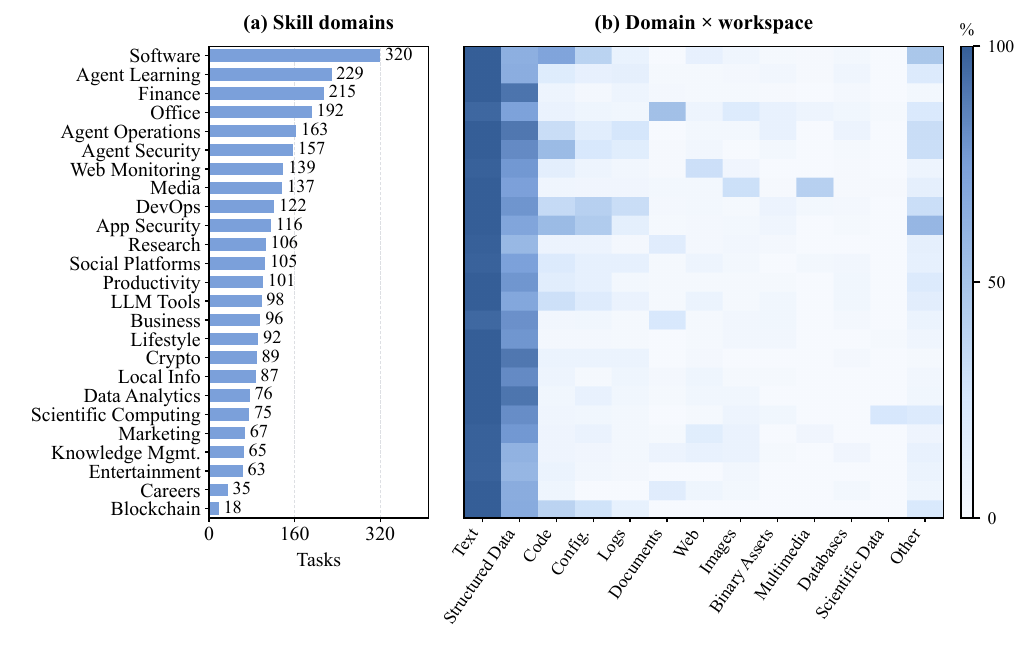}
    \caption{\textbf{Diversity of 2,963 synthesized tasks.} (a) Task counts across 25 source-skill domains, shown with abbreviated labels. (b) Percentage of tasks in each domain containing each workspace asset family; rows align with (a).}
    \label{fig:environment-diversity}
\end{figure}

Figure~\ref{fig:environment-diversity} summarizes source-skill domain coverage and its relationship to initial workspace assets.

\paragraph{Skill-domain coverage.}
Each task inherits one primary domain label from its source skill under a model-generated taxonomy. The collection spans 25 domains. The largest domain is software engineering and code quality, containing 320 tasks (10.8\%). Other prominent domains include agent memory and self-improvement, financial markets, office documents, scientific computing, media processing, and everyday applications.

\paragraph{Workspace diversity and domain--workspace combinations.}
Initial workspaces contain 46,295 files, covering 113 recognized extensions across 12 asset families and 812 distinct nonempty extension sets. These families include source code, structured data, text and markup, office documents, images, audio/video, and scientific assets. Of the 2,963 tasks, 2,707 (91.4\%) contain at least two recognized extensions, while 254 contain one and two contain none. File types are identified by their final extensions, excluding configured skill mounts, caches, and dependency directories.

The joint distribution shows domain-specific combinations: office documents and PDFs occur in 52.1\% of office-domain tasks, audio/video assets in 40.9\% of media tasks, and geospatial/scientific assets in 24.0\% of scientific-computing tasks. Thus, domain coverage is accompanied by varied workspace materials within domains.

\section{Per-Benchmark Results Across Iterations}
\label{app:iteration-results}

\begin{figure}[H]
    \centering
    \includegraphics[width=\linewidth]{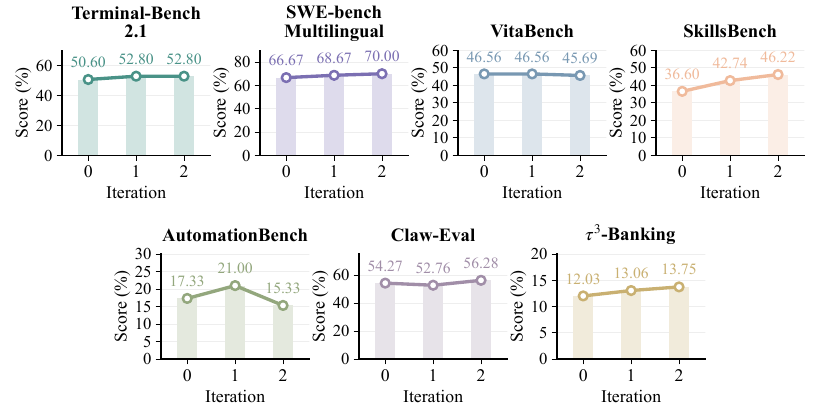}
    \caption{\textbf{Per-benchmark scores across hardening iterations.} Bars show scores and lines connect successive iterations. Labels report the scores to two decimal places.}
    \label{fig:iteration-benchmark-scores}
\end{figure}

Figure~\ref{fig:iteration-benchmark-scores} provides the individual benchmark scores underlying the average in Figure~\ref{fig:task-hardening-summary}(d). The seven-benchmark unweighted mean is 40.58, 42.51, and 42.87 at iterations 0, 1, and 2, respectively.

\section{Docling Case Study: From a Skill to a Reconciliation Environment}
\label{app:docling-case}

This case shows how a document-processing skill becomes a quarterly vendor-reconciliation environment, and how execution evidence guides two rounds of task hardening. We present the complete user request, finance brief, and setup script, followed by the blueprint's requirements, challenges, information boundaries, reference facts, and evaluation criteria. The task materials define what the solver receives; the blueprint and evaluation criteria describe the author-side contract.

\begingroup
\raggedbottom
\setlength{\parindent}{0pt}
\setlength{\parskip}{3pt}

\begin{s2ecasebox}{Task setting: complete user request}
I need your help closing out a vendor engagement for our Q3 2024 books. I'm the finance analyst at Meridian Field Institute; our finance manager has sent a close-out request for the consultancy engagement \textbf{MSA-2024-HV07} with Halden \& Verweij Consultancy BV.

The engagement file is in this workspace:

\begin{itemize}\setlength{\itemsep}{2pt}

\item \path{correspondence/2024-10-03_finance_closeout_request.md} --- the close-out brief. \textbf{This is the authoritative specification}: it defines the scope, the checks to perform, and the exact structure of the three deliverables.

\item \path{correspondence/2024-07-02_vendor_rate_increase_notice.md} --- a rate notice from the vendor; assess it against the contract before relying on it.

\item \path{contract/MSA-2024-HV07.docx} --- the consultancy agreement (fees, caps, travel policy, VAT, corrections).

\item \path{vendor_rate_card_annexB.html} --- the contractual Annex B rate card (HTML export).

\item \path{invoices/INV-2024-0731.pdf} and \path{invoices/INV-2024-0930.pdf} --- the Q3 2024 invoices.

\item \path{receipts/} --- scanned travel receipts for the quarter.

\item \path{ledger/2024_q3_payments.xlsx} --- this quarter's outgoing payments.

\item \path{archive/} --- material from a closed 2023 engagement, attached for reference only and out of scope for the 2024 reconciliation.

\end{itemize}

Please reconcile the vendor's Q3 2024 billing against the contract terms: check every billed line against the agreement and Annex B, check travel claims against the receipts, verify invoice arithmetic and VAT, and cross-check settlement against the payments ledger. The source documents are in different formats (DOCX, PDF, HTML, scanned images, XLSX) --- extract them reliably and watch that you do not lose table structure or receipt values in conversion.

Produce exactly the three deliverables specified in the brief, at these paths:

\begin{enumerate}\setlength{\itemsep}{2pt}

\item \path{reconciliation/line_items.csv}

\item \path{reconciliation/summary.json}

\item \path{reconciliation/findings.md}

\end{enumerate}

All amounts are EUR. Where the brief specifies a schema or format, follow it exactly --- the files feed our audit tooling. If your checks surface corrections, quantify them per the contract and recommend the resulting recovery action.

\end{s2ecasebox}

\begin{s2ecasebox}{Execution environment: the skill, tools, and initial workspace}
The Docling skill supplies document-conversion and OCR guidance. The task starts with \textbf{10 files in six formats}; the three requested output files do not yet exist.
\smallskip

\renewcommand{\arraystretch}{1.14}
\begin{tabularx}{\linewidth}{@{}p{0.29\linewidth}>{\raggedright\arraybackslash}p{0.13\linewidth}X@{}}
\toprule
\textbf{Initial material} & \textbf{Files} & \textbf{Role in the task} \\
\midrule
Contract and Annex B & DOCX + HTML & Rates, quarterly day cap, travel rules, VAT. \\
July/September invoices & 2 PDF & Seven billed rows to extract and reconcile. \\
Taxi and rail receipts & 2 PNG & Scanned evidence of actual travel costs. \\
Payment ledger & 1 XLSX & Establishes that both valid invoices were paid. \\
Finance brief and rate notice & 2 MD & Output schemas; a rate change effective in 2025. \\
Archived invoice & 1 PDF & Similar-looking material outside the 2024 engagement. \\
\bottomrule
\end{tabularx}
\end{s2ecasebox}
\subsection{The Complete Finance Brief}
\label{app:docling-brief}

\begin{s2ecasebox}{Workspace document: finance close-out request}
\par\medskip\textbf{Close-out request: Halden \& Verweij (MSA-2024-HV07) - Q3 2024}\par

\textbf{From:} Marieke van Dijk, Finance Manager \textbf{To:} Finance analyst (vendor close-out) \textbf{Date:} 2024-10-03

Hi,

The Q3 2024 books are closing and internal audit wants the Halden \& Verweij engagement reconciled before we release the file. Both Q3 invoices have already been paid in full (see \path{ledger/2024_q3_payments.xlsx}), so if the reconciliation surfaces corrections we will need a quantified credit-note request.

\par\medskip\textbf{Scope}\par

\begin{itemize}\setlength{\itemsep}{2pt}

\item Engagement \textbf{MSA-2024-HV07}, calendar quarter \textbf{Q3 2024} (1 July - 30 September 2024) only.

\item In-scope invoices: \path{invoices/INV-2024-0731.pdf} and \path{invoices/INV-2024-0930.pdf}.

\item Contract terms: \path{contract/MSA-2024-HV07.docx}; the contractual Annex B rate card is also exported at \path{vendor_rate_card_annexB.html}.

\item Travel receipts for the quarter are scanned in \path{receipts/}.

\item \path{archive/INV-2023-1215.pdf} was attached by ops for layout reference only - it belongs to a \textbf{closed 2023 engagement and is out of scope}. Do not include it in 2024 figures.

\item The vendor's rate notice (\path{correspondence/2024-07-02_vendor_rate_increase_notice.md}) is background only; check it against the contract before relying on it.

\end{itemize}

\par\medskip\textbf{Checks}\par

\begin{enumerate}\setlength{\itemsep}{2pt}

\item Every billed line against the contract terms (rates, day cap, travel policy, VAT).

\item Travel claims against the scanned receipts.

\item Invoice arithmetic (line amounts, subtotals, VAT, totals).

\item Settlement against the payments ledger.

\end{enumerate}

\par\medskip\textbf{Deliverables - write exactly these three files}\par

\par\medskip\textbf{1. \path{reconciliation/line_items.csv}}\par

One row per billed line, exact header:

\begin{lstlisting}[style=s2ecasecode]
invoice_id,invoice_date,description,category,quantity,unit,unit_price_eur,amount_eur_excl_vat,source_file
\end{lstlisting}

\begin{itemize}\setlength{\itemsep}{2pt}

\item \path{category} is one of \path{senior_consulting}, \path{data_analysis}, \path{mileage}, \path{taxi}, \path{rail}.

\item \path{unit} is one of \path{day}, \path{km}, \path{trip}.

\item \path{invoice_date} is ISO \path{yyyy-mm-dd}.

\item All amounts are EUR excl. 21\% VAT, 2 decimal places, no thousands separators or currency symbols.

\item Lines are recorded \textbf{as billed} on the invoices (do not yet net out corrections); \path{source_file} is the workspace-relative path of the source document.

\end{itemize}

\par\medskip\textbf{2. \path{reconciliation/summary.json}}\par

Exact structure:

\begin{lstlisting}[style=s2ecasecode]
{
  "vendor": "Halden & Verweij Consultancy BV",
  "engagement": "MSA-2024-HV07",
  "quarter": "2024-Q3",
  "invoices": [
    {"invoice_id": "...", "invoice_date": "...", "subtotal_excl_vat": 0.00, "vat": 0.00, "total_incl_vat": 0.00}
  ],
  "totals": {
    "billed_subtotal_excl_vat": 0.00, "billed_vat": 0.00, "billed_total_incl_vat": 0.00,
    "paid_total_incl_vat": 0.00,
    "corrected_subtotal_excl_vat": 0.00, "corrected_vat": 0.00, "corrected_total_incl_vat": 0.00,
    "credit_due_incl_vat": 0.00
  },
  "days": {"billed_days": 0, "cap_days": 24},
  "corrections": [
    {"invoice_id": "...", "reason": "rate_overbilling", "amount_excl_vat": 0.00}
  ]
}
\end{lstlisting}

\begin{itemize}\setlength{\itemsep}{2pt}

\item Money fields are numbers rounded to 2 decimals (round half up).

\item \path{paid_total_incl_vat} comes from the ledger (this vendor only).

\item \path{corrections} uses adjustment codes \path{rate_overbilling}, \path{travel_overclaim}, or \path{day_cap_excess}; \path{amount_excl_vat} is the positive reduction per code and invoice. Empty list if no corrections.

\item \path{corrected_*} are the amounts that \emph{should} have been billed after applying all corrections. \path{credit_due_incl_vat} is what we recover (0.00 if none).

\end{itemize}

\par\medskip\textbf{3. \path{reconciliation/findings.md}}\par

Short audit memo to me: engagement summary, each discrepancy with evidence (source document and figures), the net correction, and the recommended action (credit note amount incl. VAT, if any). Keep it under a page if you can.

Thanks - questions before the payment-run sign-off, please.

Marieke

\end{s2ecasebox}

\subsection{The Complete Environment Setup}
\label{app:docling-setup}

\begin{s2ecasebox}{Execution substrate: setup.sh}
\begin{lstlisting}[style=s2ecasecode,language=bash]
#!/usr/bin/env bash
# Environment provisioning for the docling close-out reconciliation task.
# Idempotent, non-interactive. Installs the docling CLI (skill capability) with
# CPU-only torch and warms the model cache.
set -u

PIP="pip install --no-input"

# 1) CPU-only torch + torchvision wheels (avoids multi-GB CUDA wheels).
$PIP "torch==2.13.0+cpu" "torchvision==0.28.0+cpu" \
  --index-url https://download.pytorch.org/whl/cpu

# 2) docling CLI (pinned; pulls docling-slim / docling-ibm-models / transformers).
$PIP "docling==2.123.1"

# 3) Guard: docling's dependency resolver may have overwritten torchvision with
#    the PyPI (CUDA) build whose ops fail on CPU-only boxes. Restore CPU builds
#    if the import check fails or the local +cpu tag is missing.
python3 - <<'EOF' || REINSTALL=1
import torch, torchvision
assert "+cpu" in torch.__version__ + torchvision.__version__, "non-cpu build"
EOF
if [ "${REINSTALL:-0}" = "1" ]; then
  $PIP --force-reinstall --no-deps "torch==2.13.0+cpu" "torchvision==0.28.0+cpu" \
    --index-url https://download.pytorch.org/whl/cpu
fi

# 4) Warm the HuggingFace/docling model cache with one PDF and one PNG run so the
#    solver's first conversions are fast and do not depend on network reliability.
WARM="$(mktemp -d)"
python3 - "$WARM" <<'EOF'
import sys
warm = sys.argv[1]
from reportlab.pdfgen import canvas
c = canvas.Canvas(f"{warm}/warm.pdf")
c.drawString(72, 720, "warmup invoice total 1.00")
c.save()
from PIL import Image, ImageDraw
img = Image.new("RGB", (400, 120), "white")
ImageDraw.Draw(img).text((10, 10), "WARMUP RECEIPT TOTAL EUR 1.00", fill="black")
img.save(f"{warm}/warm.png")
EOF
docling "$WARM/warm.pdf" --output "$WARM/out" >/dev/null 2>&1 || true
docling "$WARM/warm.png" --from image --output "$WARM/out" >/dev/null 2>&1 || true
rm -rf "$WARM"

# 5) Self-check (diagnostic lines).
docling --version | head -2
pip freeze 2>/dev/null | grep -Ei '^(docling|torch|torchvision)==' || true
\end{lstlisting}

\end{s2ecasebox}

\subsection{How the Blueprint Connects Requirements and Challenges}
\label{app:docling-blueprint}

The requirements specify the outputs to check. The challenges describe intended
obstacles and the behavior that would address them. The six blueprint challenges
instantiate combinations of the reusable difficulty patterns in Appendix~\ref{app:difficulty-patterns}.
Table~\ref{tab:docling-pattern-mapping} summarizes this mapping.

\begin{table}[H]
\centering
\small
\caption{Mapping from blueprint challenges in the Docling case study to the
instantiated difficulty patterns in Appendix~B.}
\label{tab:docling-pattern-mapping}
\begin{tabularx}{\linewidth}{@{}lX@{}}
\toprule
Blueprint challenge & Instantiated difficulty patterns \\
\midrule
H1 & S01 Capability routing; S06 Parameter binding; S08 Invocation context \\
H2 & U02 Distributed evidence \\
H3 & U01 Target discrimination; U06 Authority resolution; U07 Temporal validity \\
H4 & S15 Precision control; S16 Rule realization \\
H5 & L07 Cross-format agreement; L09 Aggregate agreement; L10 Claim-evidence agreement \\
H6 & S13 Schema mismatches \\
\bottomrule
\end{tabularx}
\end{table}

\begin{s2ecasebox}{Blueprint: requirements (R1--R5)}
\textbf{R1.} reconciliation/line\_items.csv contains exactly the 7 billed lines from the two Q3 2024 invoices with the brief's exact column schema and correct quantities, unit prices, and amounts (excl. VAT); archive invoice excluded

\textbf{Visible basis.} brief schema + invoices

\textbf{Evaluation.} deterministic

\textbf{Type.} objective

\textbf{Preserve during implementation.} Yes

\textbf{R2.} reconciliation/summary.json conforms to the brief's schema with correct as-billed invoice/Q3 totals (per-invoice 4989.00/1047.69/6036.69 and 5165.40/1084.73/6250.13; billed 10154.40/2132.42/12286.82; paid 12286.82) and day counts (26 billed, 24 cap)

\textbf{Visible basis.} brief schema + invoices + ledger

\textbf{Evaluation.} deterministic

\textbf{Type.} objective

\textbf{Preserve during implementation.} Yes

\textbf{R3.} summary.json corrections itemize all three discrepancies with correct ex-VAT amounts (240.00 rate over-billing, 13.00 taxi over-claim, 570.00 day-cap excess) attributed to INV-2024-0930

\textbf{Visible basis.} contract rules + invoices + receipts + brief

\textbf{Evaluation.} deterministic

\textbf{Type.} objective

\textbf{Preserve during implementation.} Yes

\textbf{R4.} summary.json corrected totals and recovery: corrected\_subtotal\_excl\_vat 9331.40, corrected\_vat 1959.59, corrected\_total\_incl\_vat 11290.99, credit\_due\_incl\_vat 995.83

\textbf{Visible basis.} brief schema + derived per contract rules

\textbf{Evaluation.} deterministic

\textbf{Type.} objective

\textbf{Preserve during implementation.} Yes

\textbf{R5.} reconciliation/findings.md is an audit-ready memo: identifies all three discrepancies with evidence (invoice lines, contract clauses, receipt amounts), explains why the 2025 rate notice does not apply to Q3 2024, and recommends a credit note of EUR 995.83 incl. VAT

\textbf{Visible basis.} brief + all workspace evidence

\textbf{Evaluation.} llm judge

\textbf{Type.} subjective

\textbf{Preserve during implementation.} Yes

\end{s2ecasebox}

\begin{s2ecasebox}{Blueprint challenge H1: multi-format tool use and extraction fidelity}
\textbf{Plausible failure.} Weak solver hand-transcribes PDFs/images, loses the bordered table structure, or never OCRs the receipts, so line amounts or the taxi receipt value are wrong/missing

\textbf{Strong solution behavior.} Uses docling per format into a controlled output dir (PDF/DOCX/HTML/image OCR), cross-checks extracted tables against line arithmetic

\end{s2ecasebox}

\begin{s2ecasebox}{Blueprint challenge H2: distributed rules across documents}
\textbf{Plausible failure.} Applies only the most salient rule (e.g. VAT) and never notices the day cap or taxi cap buried in contract clauses

\textbf{Strong solution behavior.} Extracts the full rule set (rates, 24-day aggregate cap with lowest-rate-first deduction, taxi at-cost capped at 40, VAT 21\%) and evaluates every line against every applicable rule

\end{s2ecasebox}

\begin{s2ecasebox}{Blueprint challenge H3: target-object identification / scoping}
\textbf{Plausible failure.} Includes archive/INV-2023-1215.pdf in the 2024 totals, or treats the vendor's 2025 rate notice as applicable and 'validates' the EUR 450 rate

\textbf{Strong solution behavior.} Scopes to MSA-2024-HV07 Q3 2024 per the brief; reads the notice's effective date (2025-01-01) and the contract's Annex B governing clause, so the 450 rate is flagged as over-billing, not accepted

\end{s2ecasebox}

\begin{s2ecasebox}{Blueprint challenge H4: rule realization and numerical precision}
\textbf{Plausible failure.} Computes VAT with float drift or inconsistent rounding, mis-orders cap deductions (deducts senior days), or double-counts the taxi cap vs receipt, producing a wrong credit figure

\textbf{Strong solution behavior.} Works in cents/Decimal with round-half-up 2dp; payable taxi = min(52.00 claim, 39.00 receipt, 40.00 cap) = 39.00; 2 excess analyst days x 285.00; credit = 823.00 + 172.83 VAT = 995.83

\end{s2ecasebox}

\begin{s2ecasebox}{Blueprint challenge H5: cross-artifact consistency}
\textbf{Plausible failure.} Findings.md and summary.json disagree (memo says one credit amount, JSON another) because corrections were computed ad-hoc per artifact

\textbf{Strong solution behavior.} A single consistent correction set flows into CSV totals, JSON corrections/totals, and the memo

\end{s2ecasebox}

\begin{s2ecasebox}{Blueprint challenge H6: schema fidelity under an explicit contract}
\textbf{Plausible failure.} Invents column names / JSON keys instead of following the brief, breaking downstream audit tooling

\textbf{Strong solution behavior.} Maps the extracted invoice and contract information into the exact CSV and JSON target schemas specified in the finance brief, preserving field meanings and relationships.

\end{s2ecasebox}

\subsection{Information Boundaries and Reference Facts}
\label{app:docling-boundaries}

\begin{s2ecasebox}{Blueprint: information boundaries}
\textbf{Explicit in the task instruction.}
\begin{itemize}\setlength{\itemsep}{2pt}
\item The overall objective (Q3 2024 close-out reconciliation for vendor Halden \& Verweij under MSA-2024-HV07)
\item The three required deliverables and their exact root-relative paths (reconciliation/line\_items.csv, reconciliation/summary.json, reconciliation/findings.md)
\item That the finance brief (correspondence/2024-10-03\_finance\_closeout\_request.md) is the authoritative specification of deliverable schemas, scope, and the checks to perform
\item That amounts are EUR, VAT is handled per the contract, and archive/ material is out of scope
\end{itemize}

\textbf{Present in the workspace.}
\begin{itemize}\setlength{\itemsep}{2pt}
\item All commercial rules (rates, caps, VAT 21\%, cap-deduction ordering, effective-date rule) in contract and rate card, NOT restated in the prompt
\item The exact deliverable schemas in the finance brief
\item The receipt amounts (39.00 taxi, 88.40 rail) only inside the scanned PNGs
\item The 2025 rate-increase notice as a plausibly-confusing but non-applicable document
\end{itemize}

\textbf{Discovered through execution.}
\begin{itemize}\setlength{\itemsep}{2pt}
\item That docling is the right tool and which --from/--output flags each format needs
\item That the archive invoice shares structure with the in-scope invoices but must be excluded
\item That the vendor over-billing stems from early application of the announced 2025 rate
\end{itemize}

\textbf{Inferred from the materials.}
\begin{itemize}\setlength{\itemsep}{2pt}
\item payable taxi amount = min(claimed, receipted, cap) from the at-cost + cap rule
\item cap excess (2 days) falls on analyst lines per the lowest-rate-first rule
\end{itemize}

\textbf{No hidden assumptions.}
\begin{itemize}\setlength{\itemsep}{2pt}
\item No fact requires information outside the workspace; VAT rate, caps, rates, and rounding expectations are all stated in workspace documents
\item The 2025 rate notice explicitly states its effective date so its non-applicability is resolvable from context, not guessing
\end{itemize}

\end{s2ecasebox}

\begin{s2ecasebox}{Blueprint: fixed evidence and external access}
\textbf{Search expected.} No

\textbf{Temporal mode.} fixed

\begin{itemize}\setlength{\itemsep}{2pt}
\item All amounts, dates, VAT rate, caps, and rates are fixed workspace facts as of the scenario date 2024-10-03
\item Credit note EUR 995.83 incl. VAT is derivable from workspace inputs alone
\end{itemize}
\textbf{Evaluation strategy.} Fully offline evaluation against fixed reference facts, using deterministic checks for objective criteria and an LLM judge for subjective criterion.

\end{s2ecasebox}

\begin{s2ecasebox}{Blueprint: complete reference facts}
\begin{lstlisting}[style=s2ecasecode]
{
  "inv1_subtotal_excl_vat": 4989.0,
  "inv1_vat": 1047.69,
  "inv1_total_incl_vat": 6036.69,
  "inv2_subtotal_excl_vat": 5165.4,
  "inv2_vat": 1084.73,
  "inv2_total_incl_vat": 6250.13,
  "billed_subtotal_excl_vat": 10154.4,
  "billed_vat": 2132.42,
  "billed_total_incl_vat": 12286.82,
  "paid_total_incl_vat": 12286.82,
  "billed_days": 26,
  "cap_days": 24,
  "excess_days": 2,
  "correction_rate_overbilling_excl_vat": 240.0,
  "correction_taxi_overclaim_excl_vat": 13.0,
  "correction_day_cap_excl_vat": 570.0,
  "corrections_total_excl_vat": 823.0,
  "corrections_vat": 172.83,
  "corrected_subtotal_excl_vat": 9331.4,
  "corrected_vat": 1959.59,
  "corrected_total_incl_vat": 11290.99,
  "credit_due_incl_vat": 995.83,
  "line_item_count": 7
}
\end{lstlisting}

\end{s2ecasebox}

\begin{s2ecasebox}{Blueprint: derivations of the reference outcomes}
\textbf{INV-2024-0731 totals 4989.00/1047.69/6036.69.} 9x420.00 + 4x285.00 + 300x0.23 = 3780+1140+69 = 4989.00; VAT 4989.00x0.21 = 1047.69 exact; total 6036.69

\textbf{Validation.} calculation + docling extraction cross-check

\textbf{INV-2024-0930 billed totals 5165.40/1084.73/6250.13.} 8x450.00 + 5x285.00 + 52.00 + 88.40 = 5165.40; VAT 1084.734 -> 1084.73 (2dp half-up as printed); total 6250.13

\textbf{Validation.} calculation + docling extraction cross-check

\textbf{rate over-billing 240.00.} Annex B senior rate 420.00 governs Q3 2024; 450.00 applies only from 2025-01-01; the invoice bills 8 senior days at 450.00; 8x(450-420)=240.00

\textbf{Validation.} calculation

\textbf{taxi over-claim 13.00.} travel reimbursed at cost with receipt, taxi capped at 40.00 per trip; receipt = 39.00; payable = min(52.00, 39.00, 40.00) = 39.00; over-claim 13.00

\textbf{Validation.} OCR extraction + calculation

\textbf{day-cap correction 570.00, excess 2 days.} Q3 billed days 9+4+8+5 = 26 vs cap 24; excess 2 days deducted lowest-rate-first from the final invoice of the quarter -> 2 analyst days x 285.00 = 570.00

\textbf{Validation.} calculation

\textbf{credit\_due\_incl\_vat 995.83 and corrected totals.} corrections 823.00 excl VAT; VAT on corrections 172.83; corrected 9331.40/1959.59/11290.99; both invoices fully paid (12286.82); credit = 12286.82 - 11290.99 = 995.83

\textbf{Validation.} calculation + ledger cross-check

\end{s2ecasebox}

\begin{s2ecasebox}{Blueprint: invariants and critical requirements}
\begin{itemize}\setlength{\itemsep}{2pt}
\item line\_items.csv has exactly 7 data rows; every amount equals quantity x unit\_price to the cent
\item summary.json internal consistency: invoice totals sum to billed totals; corrected + corrections = billed (excl VAT); credit\_due = paid - corrected\_total
\item No line references archive/INV-2023-1215.pdf and the 2025 rate is not treated as applicable to Q3 2024
\end{itemize}
\textbf{Requirements carrying the central challenges:} R3 and R4.

\textbf{Critical failure requirements:} R3 and R4.

\end{s2ecasebox}

\subsection{Complete Evaluation Criteria}
\label{app:docling-evaluation}

\begin{s2ecasebox}{Rubric: line items csv}
\textbf{Criterion.} reconciliation/line\_items.csv follows the brief's exact 9-column schema and contains exactly the 7 billed Q3 2024 lines with correct invoice dates, quantities, units, unit prices and ex-VAT amounts, excluding the out-of-scope 2023 archive invoice.

\textbf{Weight.} 0.25

\textbf{Evaluation method.} deterministic

\end{s2ecasebox}

\begin{s2ecasebox}{Rubric: summary billed totals}
\textbf{Criterion.} reconciliation/summary.json conforms to the brief's schema: both invoices with their as-billed subtotal/VAT/total (4,989.00/1,047.69/6,036.69 and 5,165.40/1,084.73/6,250.13), quarterly billed totals (10,154.40 / 2,132.42 / 12,286.82), ledger-paid total 12,286.82, and day counts 26 billed vs 24 cap.

\textbf{Weight.} 0.15

\textbf{Evaluation method.} deterministic

\end{s2ecasebox}

\begin{s2ecasebox}{Rubric: corrections breakdown}
\textbf{Criterion.} summary.json corrections itemize all three contract violations with correct ex-VAT amounts, attributed to INV-2024-0930: rate\_overbilling 240.00 (8 senior days at 450 vs the Annex B rate 420 that governs 2024), travel\_overclaim 13.00 (52.00 taxi claim vs 39.00 receipt under the 40.00 cap), day\_cap\_excess 570.00 (26 vs 24 billed days, 2 analyst days at 285.00 deducted lowest-rate-first).

\textbf{Weight.} 0.20

\textbf{Evaluation method.} deterministic

\end{s2ecasebox}

\begin{s2ecasebox}{Rubric: corrected totals credit}
\textbf{Criterion.} summary.json reports the corrected position: corrected\_subtotal\_excl\_vat 9,331.40, corrected\_vat 1,959.59, corrected\_total\_incl\_vat 11,290.99, and credit\_due\_incl\_vat 995.83 (both invoices were paid in full, so this is the recoverable amount).

\textbf{Weight.} 0.15

\textbf{Evaluation method.} deterministic

\end{s2ecasebox}

\begin{s2ecasebox}{Rubric: findings memo}
\textbf{Criterion.} reconciliation/findings.md is an evidence-grounded audit memo to the finance manager covering all discrepancies and the recommended recovery.

\textbf{Weight.} 0.25

\textbf{Evaluation method.} llm judge

\textbf{Judge instruction.} Score from 0 to 1 the memo reconciliation/findings.md against these criteria. Ground truth (do not reveal beyond judging): (1) INV-2024-0930 bills 8 senior consultant days at EUR 450.00/day although Annex B of MSA-2024-HV07 (governing all 2024 work) sets EUR 420.00/day; the vendor's rate notice raises rates only from 2025-01-01, so the over-billing is 8 x 30 = EUR 240.00 excl. VAT. (2) The 12 Sep taxi is claimed at EUR 52.00 but the receipt (receipts/2024-09-12\_taxi\_citycab.png) shows EUR 39.00 and the contract caps taxi at EUR 40.00/trip and reimburses at receipted cost, so EUR 13.00 is over-claimed; the EUR 88.40 rail claim matches its receipt. (3) Q3 billed days total 26 (13 July + 13 September) vs the contract's aggregate cap of 24 per quarter; the 2 excess days are deducted from the lowest day-rate lines (data analyst, EUR 285.00) on the final invoice of the quarter, i.e. EUR 570.00 excl. VAT. Total corrections EUR 823.00 excl. VAT; with 21\% VAT the credit note due is EUR 995.83 incl. VAT; corrected totals 9,331.40 / 1,959.59 / 11,290.99; both invoices were already paid in full (12,286.82), so the amount must be recovered by credit note. Award roughly: 0.35 for identifying and correctly quantifying all three discrepancies (0.12-ish each; a discrepancy counts only if its cause AND amount are right), 0.25 for the correct net correction and credit-note recommendation (EUR 995.83 incl. VAT) explicitly tied to the fact that both invoices were fully paid, 0.15 for correctly explaining why the 2025 rate notice does not legitimize the 450 rate (effective 2025-01-01; Annex B governs 2024), 0.15 for evidence grounding (memo cites the specific source documents/figures: invoice lines, contract clauses or Annex B, receipt values, ledger payments) without fabricated content, 0.10 for clarity and audit-readiness (addressed to the finance manager, concise, internally consistent figures). Deduct for any figure that contradicts the ground truth (e.g. a different credit amount), for inventing findings not supported by the documents, or if the memo conflicts with reconciliation/summary.json when that file is also available. An otherwise strong memo in different wording or structure must still score highly.

\textbf{Files available to the judge.} reconciliation/findings.md, reconciliation/summary.json, reconciliation/line\_items.csv

\end{s2ecasebox}

\begin{s2ecasebox}{Rubric: passing condition}
\textbf{Original evaluator pass threshold.} 0.7

A passing solution converts the mixed-format close-out package into the exact-schema line-item CSV and summary JSON, finds and correctly quantifies all three contract violations (240.00 + 13.00 + 570.00 excl. VAT), derives the corrected totals and the EUR 995.83 credit note, and explains them in an evidence-grounded audit memo.
\end{s2ecasebox}

\subsection{How Execution Evidence Informed Two Hardening Revisions}
\label{app:docling-hardening}
\paragraph{Consistency validation.}
Before using rollout evidence for hardening, the task conditions and rubric
results were checked against the blueprint. The workspace contained the
required evidence, the evaluator criteria were consistent with the reference
facts, and no evaluator or environment inconsistency affecting the diagnosis
was identified. The validated execution evidence was then used for diagnosis.
\begin{s2ecasebox}{First hardening: extend a workflow the solver already completed}
\textbf{Observed execution.}
The original rollout received full credit on all five rubric items. The diagnosis agent read that result and inspected execution records showing document conversion, deliverable writes, and numerical self-checks. The complete self-check command was:
\begin{lstlisting}[style=s2ecasecode,language=bash]
cd /app && python3 -c "
import json, csv
# validate json
d=json.load(open('reconciliation/summary.json'))
t=d['totals']
print('invoices sum subtotal:', round(sum(i['subtotal_excl_vat'] for i in d['invoices']),2))
print('invoices sum vat:', round(sum(i['vat'] for i in d['invoices']),2))
print('invoices sum total:', round(sum(i['total_incl_vat'] for i in d['invoices']),2))
print('billed totals consistent:', t['billed_subtotal_excl_vat']==10154.40 and t['billed_vat']==2132.42 and t['billed_total_incl_vat']==12286.82)
# corrected
sub=4989.00+4342.40
vat=1047.69+round(4342.40*0.21,2)
tot=6036.69+round(4342.40+4342.40*0.21,2)
print('corrected sub/vat/tot',sub,vat,tot)
print('credit',12286.82-round(tot,2))
print('corrections sum',sum(c['amount_excl_vat'] for c in d['corrections']))
# csv rows
r=list(csv.DictReader(open('reconciliation/line_items.csv')))
print('csv rows',len(r))
print('csv subtotal',round(sum(float(x['amount_eur_excl_vat']) for x in r),2))
# verify each invoice line total
for x in r:
    assert abs(float(x['amount_eur_excl_vat'])-round(float(x['quantity'])*float(x['unit_price_eur']),2))<0.01
print('line arithmetics OK')
# days
print('days',sum(int(x['quantity']) for x in r if x['category'] in ('senior_consulting','data_analysis')))
"
\end{lstlisting}
\textbf{Design response.}
The diagnosis agent judged the current task insufficiently challenging. New requirements make previously straightforward checks consequential:
\begin{itemize}
\setlength{\itemsep}{1pt}
\item Put the principal-consultant rate only in the signed DOCX amendment: the HTML rate card alone is insufficient.
\item Rebill one rail journey across months: individual invoice checks must become a cross-invoice comparison.
\item Add another engagement of the same vendor, with its own ledger payment: scope must be checked in both invoices and payments.
\item Print EUR 1,435 for $5\times285=\text{EUR }1,425$: preserve the original billed line and separately record the EUR 10 correction.
\item Partially pay an invoice: distinguish billed amounts from cash already paid.
\end{itemize}
This is hardening informed by a \emph{successful} rollout. The new duplicate-claim and partial-payment challenges were proposed extensions.
\end{s2ecasebox}

\begin{s2ecasebox}{Second hardening: make document extraction robustness consequential}
\textbf{Observed execution.}
A later archived rollout shows four aborted OCR conversions caused by an incorrectly split output path.
The command passed \texttt{/tmp/} and \texttt{receipts\_out} as separate arguments after \texttt{--output}, causing each conversion to return \texttt{Aborted.} The solver nevertheless recovered by directly reading the PNG receipts with the image-reading tool and obtained the required receipt amounts. This recovery was valid, but it also exposed a brittle document-processing path: the intended OCR/conversion procedure failed, and the rollout succeeded only because the relevant evidence happened to remain directly readable through an alternative interface.

\textbf{Design response.}
To make this weakness consequential, the hardened task places required evidence in a text-less scanned PDF whose content must be extracted reliably as part of the document-processing workflow.
The solver must correctly invoke the conversion/OCR procedure, preserve the extracted structure, and use the resulting evidence in the reconciliation.
This revision directly targets the brittle invocation behavior observed in the rollout. Additional revisions increase the amount and interaction of extracted evidence, including date-dependent rates, an unreceipted claim, and a day-cap deduction spanning two rate tiers.
\end{s2ecasebox}

\subsection{How the Task Changes Across Hardening}
\label{app:docling-outcome}

\begin{s2ecasebox}{What changed across the task versions}
\renewcommand{\arraystretch}{1.15}
\begin{tabularx}{\linewidth}{@{}Xrrr@{}}
\toprule
& \textbf{Original} & \textbf{Hardening 1} & \textbf{Hardening 2} \\
\midrule
Workspace files & 10 & 14 & 21 \\
Input formats & 6 & 6 & 6 \\
Valid invoices & 2 & 3 & 4 \\
Billed rows & 7 & 14 & 20 \\
Designed discrepancies & 3 & 7 & 12 \\
Partially paid invoices & 0 & 1 & 2 \\
\bottomrule
\end{tabularx}
\par\smallskip
The comparison shows that task complexity increases across hardening stages
through more records, interacting discrepancies, and additional settlement dependencies.
\end{s2ecasebox}

\endgroup

\end{document}